\documentclass[10pt,twocolumn,letterpaper]{article}

\usepackage{iccv}
\usepackage{times}
\usepackage{epsfig}
\usepackage{graphicx}
\usepackage{amsmath}
\usepackage{amssymb}

\usepackage[usenames,dvipsnames]{color}
\usepackage{makecell}
\usepackage{cuted}
\usepackage[misc]{ifsym}
\usepackage{balance}
\usepackage{tabularx}
\newcolumntype{Y}{>{\centering\arraybackslash}X}
\usepackage{amsmath,bm}
\usepackage{fnpct}

\usepackage[pagebackref=true,breaklinks=true,letterpaper=true,colorlinks,bookmarks=false]{hyperref}

\newcommand{\cavan}[1]{{\color{red}(cavan: {#1})}} % cavan comments
\newcommand{\liming}[1]{{\color{blue}{#1}}} % liming comments
\iccvfinalcopy % *** Uncomment this line for the final submission

\def\iccvPaperID{3038} % *** Enter the ICCV Paper ID here

\begin{document}

%%%%%%%%% TITLE
\title{\vspace{-0.1cm}Focal Frequency Loss for Image Reconstruction and Synthesis\vspace{-0.282cm}}

\author{Liming Jiang$^{1}$ \hspace{12pt} Bo Dai$^{1}$ \hspace{12pt} Wayne Wu$^{2}$ \hspace{12pt} Chen Change Loy$^{1\textrm{\Letter}}$\\[2pt]
$^1$S-Lab, Nanyang Technological University \hspace{12pt} $^2$SenseTime Research\\[1pt]
{\tt\small \{liming002, bo.dai, ccloy\}@ntu.edu.sg} \hspace{12pt}
{\tt\small wuwenyan@sensetime.com}\vspace{-0.15cm}
}

\maketitle
%\ificcvfinal\thispagestyle{empty}\fi

% !TEX root = ../main_arxiv.tex

\begin{abstract}
\label{sec:abstract}

Image reconstruction and synthesis have witnessed remarkable progress thanks to the development of generative models.
%Despite the remarkable progress of image reconstruction and synthesis by generative modeling using deep neural networks, gaps could still exist between the real and generated images, especially in the frequency domain.
Nonetheless, gaps could still exist between the real and generated images, especially in the frequency domain.
%
%Despite the remarkable success of generative models in creating photorealistic images using deep neural networks, gaps could still exist between the real and generated images, especially in the frequency domain.
%
In this study, we show that narrowing gaps in the frequency domain can ameliorate image reconstruction and synthesis quality further.
We propose a novel focal frequency loss, which allows a model to adaptively focus on frequency components that are hard to synthesize by down-weighting the easy ones.
This objective function is complementary to existing spatial losses, offering great impedance against the loss of important frequency information due to the inherent bias of neural networks.
We demonstrate the versatility and effectiveness of focal frequency loss to improve popular models, such as VAE, pix2pix, and SPADE, in both perceptual quality and quantitative performance. We further show its potential on StyleGAN2.\footnote{\hspace{0.07cm}GitHub: \href{https://github.com/EndlessSora/focal-frequency-loss}{https://github.com/EndlessSora/focal-frequency-loss}.}\footnote{\hspace{0.07cm}Project page: \href{https://www.mmlab-ntu.com/project/ffl/index.html}{https://www.mmlab-ntu.com/project/ffl/index.html}.}
%
%Moreover, a series of variants for focal frequency loss are carefully studied, and several practical considerations are discussed for potential further improvement.

\end{abstract}
% !TEX root = ../main_arxiv.tex

\section{Introduction}
\label{sec:introduction}

We have seen remarkable progress in image reconstruction and synthesis along with the development of generative models~\cite{ae,vae,GAN,glow,pixelcnn}, and the progress continues with the emergence of various powerful deep learning-based approaches~\cite{stylegan2,SPADE,alae,nvae}. 
Despite their immense success, one could still observe the gaps between the real and generated images in certain cases.

\if 0
Image reconstruction and synthesis have made remarkable breakthroughs with the fast development of generative models~\cite{ae,vae,GAN,glow,pixelcnn}, especially autoencoders (AE) and generative adversarial networks (GAN).
%
Existing powerful reconstruction and synthesis methods are mainly achieved using deep neural networks~\cite{stylegan2,SPADE,alae,nvae}.
%
Despite their immense success, one could still observe the gaps between the real and generated images in certain cases.
\fi

%\liming{Image reconstruction and synthesis have made remarkable breakthroughs with the rapid development of generative models~\cite{ae,vae,GAN,glow,pixelcnn}, especially autoencoders (AE) and generative adversarial networks (GAN).}
%%
%%Generative models \cite{ae,vae,GAN,glow,pixelcnn} have achieved remarkable performance in capturing high-dimensional representations of visual data and creating photorealistic images.
%%%
%Existing state-of-the-art generative models are mainly constructed using deep neural networks~\cite{stylegan2,SPADE,alae,nvae}.
%%
%Despite their immense success in \liming{image reconstruction and synthesis}, one could still observe the gaps between the real and generated images in certain cases.
%%, especially when the model ability is just moderate.

\begin{figure}[t]
	\centering
%	\vspace{-0.35cm}
	\includegraphics[width=\linewidth]{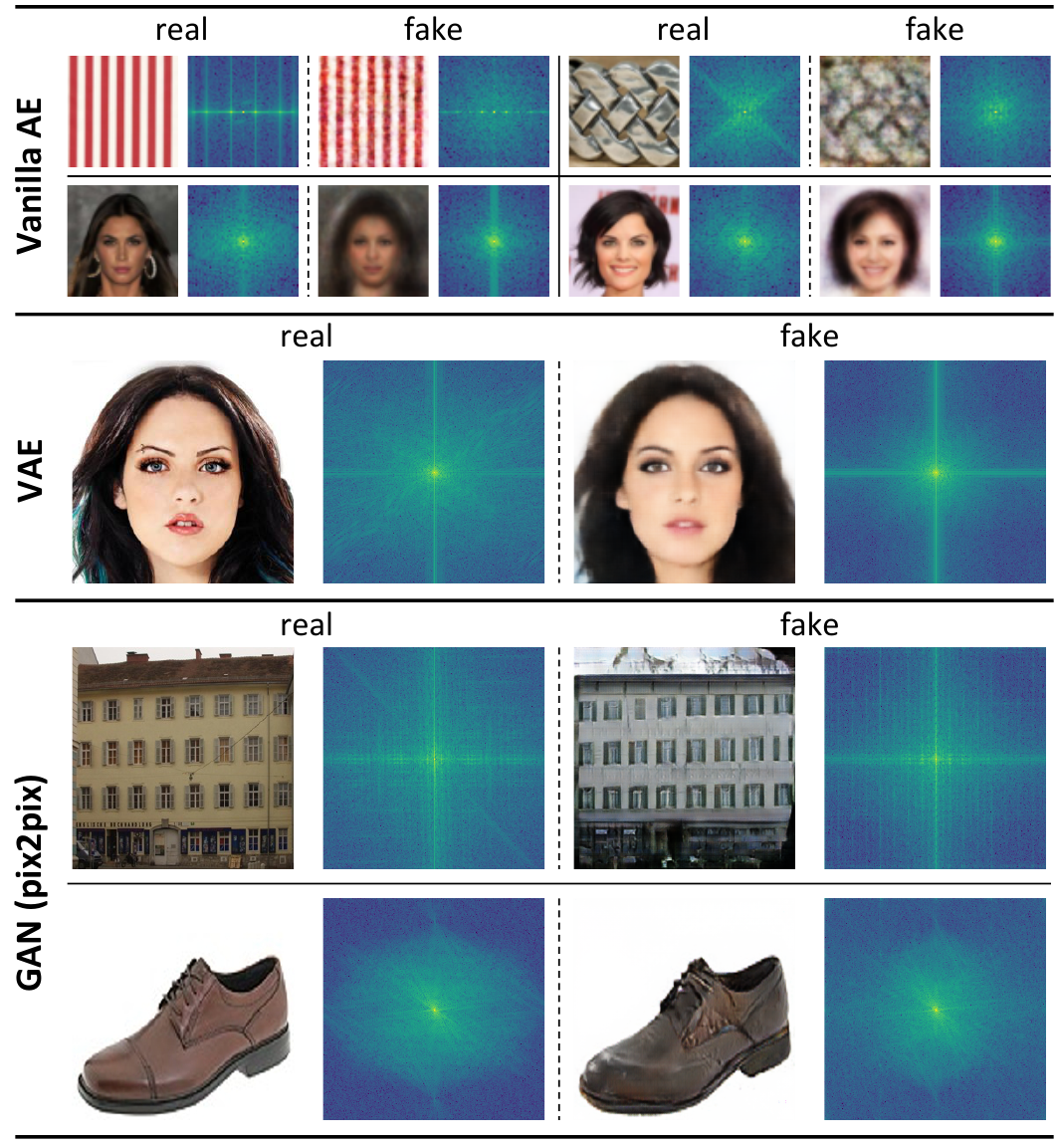}
%	\vspace{-0.5cm}
	\caption{Frequency domain gaps between the real and the generated images by typical generative models in image reconstruction and synthesis. Vanilla AE~\cite{ae} loses important frequencies, leading to blurry images (Row $1$ and $2$). VAE~\cite{vae} biases to a limited spectrum region (Row $3$), losing high-frequency information (outer regions and corners). Unnatural periodic patterns can be spotted on the spectra of images generated by GAN (pix2pix)~\cite{pix2pix} (Row $4$), consistent with the observable checkerboard artifacts (zoom in for view). In some cases, a frequency spectrum region shift occurs to GAN-generated images (Row $5$).}
	\label{fig:teaser}
	\vspace{-0.4cm}
\end{figure}

These gaps are sometimes manifested in the form of artifacts that are discernible. 
For instance, upsampling layers using transposed convolutions tend to produce checkerboard artifacts~\cite{checkerboardarti}. % , an important cue that affects the perceptual quality of synthesized images.
The gaps, in some other cases, may only be revealed through the frequency spectrum analysis. Recent studies~\cite{cnndetection,artifactsganfake,fakeretouch} in media forensics have shown some notable periodic patterns in the frequency spectra of manipulated images, which may be consistent with artifacts in the spatial domain.
In Figure~\ref{fig:teaser}, we show some paired examples of real images and the fake ones generated by typical generative models for image reconstruction and synthesis.
It is observed that the frequency domain gap between the real and fake images may be a common issue for these methods, albeit in slightly different forms.
%
%\textbf{(phenomenon and motivation)}

The observed gaps in the frequency domain may be imputed to some inherent bias of neural networks when applied to reconstruction and synthesis tasks.
%Based on the observations above, we are curious about the reason for such frequency domain gaps of generative models.
%we argue that it is critical to minimize the frequency domain differences between the real and fake images to ameliorate the synthesis quality further, especially the periodic artifacts and image details.
%
%We find that the gap may be imputed to some inherent crux of neural networks.
%However, this is \textit{non-trivial} given the inherent crux of neural networks.
%
Fourier analysis highlights a phenomenon called \textit{spectral bias}~\cite{spectralbias,nerf,fourierfeatures}, a learning bias of neural networks towards low-frequency functions. Thus, generative models tend to eschew frequency components that are hard to synthesize, \ie, hard frequencies, and converge to an inferior point.
\textit{F-Principle}~\cite{fprinciple} shows that the priority of fitting certain frequencies in a network is also different throughout the training, usually from low to high.
Consequently, it is difficult for a model to maintain important frequency information as it tends to generate frequencies with a higher priority.
%
%\textbf{(analysis and challenges)}

\if 0
In this paper, we study the frequency domain gap between real and fake images and explore ways to ameliorate reconstruction and synthesis quality by narrowing this gap.
%
We aim at devising a generic loss function that is complementary to existing spatial losses.
%
\cavan{I know Liming quite insists on this but I think we can actually remove the content the `three yardsticks' without affecting the appreciation of the idea. And it actually make the intro more concise. I would rather leave the space for including more discussion/result in the experiment. For your consideration.}
We set forth \textit{three yardsticks} when designing this loss function:
%
1) The loss form is built on a space where frequencies of an image can be well represented and distinguished, facilitating optimization in the frequency dimension.
%
2) The loss enables a network to focus on frequency components that are hard to synthesize, \ie, hard frequencies, which may be pivotal for quality improvement.
%
3) The focused frequencies should be updated dynamically as the training proceeds to bridge frequency domain gaps of different models adaptively.
\fi

%Due to the frequency bias nature, we wish to highlight that the ability of neural networks should be leveraged to the largest extent by adaptively focusing on more important frequencies that might be lost most during training.

In this paper, we carefully study the frequency domain gap between real and fake images and explore ways to ameliorate reconstruction and synthesis quality by narrowing this gap.
%
%We aim at devising a generic loss function that is complementary to existing spatial losses.
%
%To meet this objective, we carefully study the nature of frequency components in images.
%
Existing methods~\cite{vae,pix2pix,SPADE} usually adopt pixel losses in the spatial domain,
%to cope with different image generation tasks. However
while spatial domain losses hardly help a network find hard frequencies and synthesize them, in that every pixel shares the same significance for a certain frequency.
%
%In contrast, we directly take generative models into the frequency domain.
%
In comparison, we transform both the real and generated samples to their frequency representations using the standard discrete Fourier transform (DFT). The images are decomposed into sines and cosines, exhibiting periodic properties. Each coordinate value on the frequency spectrum depends on all the image pixels in the spatial domain, representing a specific spatial frequency.
Explicitly minimizing the distance of coordinate values on the real and fake spectra can help networks easily locate difficult regions on the spectrum, \ie, hard frequencies.
%
%\textbf{(Drawbacks of existing methods that cannot meet these points. Difference of our method, in line with point 1))}

To tackle these hard frequencies, inspired by hard example mining~\cite{hempartmodel,ohem} and focal loss~\cite{focalloss}, we propose a simple yet effective frequency-level objective function, named \textit{focal frequency loss}.
We map each spectrum coordinate value to a Euclidean vector in a two-dimensional space, with both the amplitude and phase information of the spatial frequency put under consideration.
The proposed loss function is defined by the scaled Euclidean distance of these vectors by down-weighting easy frequencies using a dynamic spectrum weight matrix. Intuitively, the matrix is updated on the fly according to a non-uniform distribution on the current loss of each frequency during training.
The model will then rapidly focus on hard frequencies and progressively refine the generated frequencies to improve image quality.
%
%\textbf{(After find, how to handle? Focus, focal frequency loss, brief idea, dynamic, in line with point 2) and 3))}

The main \textbf{contribution} of this work is a novel focal frequency loss that directly optimizes generative models in the frequency domain. We carefully motivate how a loss can be built on a space where frequencies of an image can be well represented and distinguished, facilitating optimization in the frequency dimension. We further explain the way that enables a model to focus on hard frequencies, which may be pivotal for quality improvement.
Extensive experiments demonstrate the effectiveness of the proposed loss on representative baselines~\cite{ae,vae,pix2pix,SPADE}, and the loss is complementary to existing spatial domain losses such as perceptual loss~\cite{perceptualloss}.
We further show the potential of focal frequency loss to improve state-of-the-art StyleGAN2~\cite{stylegan2}.

% !TEX root = ../main_arxiv.tex

\section{Related Work}
\label{sec:relatedwork}

%\vspace{0.1cm}
\noindent
\textbf{Image reconstruction and synthesis.}
%
%Recent advances~\cite{ae,vae,GAN,glow,pixelcnn} of generative models are built on deep neural networks, showing impressive capability in capturing high-level latent representations of images and synthesizing new data.
%
Autoencoders (AE) \cite{ae,vae} and generative adversarial networks (GAN)~\cite{GAN} are two popular models for image reconstruction and synthesis.
The vanilla AE~\cite{ae} aims at learning latent codes while reconstructing images. It is typically used for dimensionality reduction and feature learning. 
Autoencoders have been widely used to generate images since the development of variational autoencoders (VAE)~\cite{vae,convae}. Their applications have been extended to various tasks, \eg, face manipulation~\cite{DeepFakes,DFL,deeperforensics1,dfc20}.
GAN~\cite{GAN,congan,DCGAN}, on the other hand, is extensively applied in face generation~\cite{pggan,stylegan,stylegan2}, image-to-image translation~\cite{pix2pix,cyclegan,stargan,tsit}, style transfer~\cite{UNIT,MUNIT}, and semantic image synthesis~\cite{pix2pixhd,SPADE,CC-FPSE}.
Existing approaches usually apply spatial domain loss functions, \eg, perceptual loss~\cite{perceptualloss}, to improve quality while seldom consider optimization in the frequency domain.
%
%Existing approaches usually apply spatial domain loss functions~\cite{perceptualloss} to generate images while seldom consider improving quality via the frequency domain.
%
Spectral regularization~\cite{specreg} presents a preliminary attempt.
Different from~\cite{perceptualloss,specreg}, the proposed focal frequency loss dynamically focuses the model on hard frequencies by down-weighting the easy ones and ameliorates image quality through the frequency domain directly.
%The proposed focal frequency loss differs from and outperforms these methods \cavan{explain what is the difference. Specify which methods.}.
%
Some concurrent works on image reconstruction and synthesis via the frequency domain include~\cite{fdit,swagan,specgan}.

%\vspace{0.1cm}
\noindent
\textbf{Frequency domain analysis of neural networks.}
%
%Recent studies have been analyzing neural networks from the frequency domain aspect.
%
In addition to the studies~\cite{spectralbias,nerf,fourierfeatures,fprinciple} we discussed in the introduction, we highlight some recent works that analyze neural networks through the frequency domain.
%
%Some studies~\cite{spectralbias,nerf,fourierfeatures} employ Fourier analysis to highlight the phenomenon of spectral bias in neural networks towards learning low-frequency functions.
%
%F-principle~\cite{fprinciple} reports that networks usually fit target functions from low to high frequencies.
%
Using coordinate-based MLPs, Fourier features~\cite{fourierfeatures,randomfeatures} and positional encoding~\cite{nerf,attention} are adopted to recover missing high frequencies in single image regression problems.
Besides, several studies have incorporated frequency analysis with network compression~\cite{fasternetjpeg,learninfreq,stylewavelet,compressfreq,notsobiggan} and feature reduction~\cite{featreduction,cnnpack} to accelerate the training and inference of networks.
The application areas of the frequency domain analysis have been further extended, including media forensics~\cite{cnndetection,artifactsganfake,fakeretouch,freqdeepfakeicml}, super-resolution~\cite{freqsepsr,dmawaresr}, generalization analysis~\cite{generalizefreq,rda}, magnetic resonance imaging~\cite{jointfreq}, image rescaling~\cite{invertrescale}, \etc.
%
%\cavan{this sentence is not complete, what do these studies to with frequency domain analysis?}.
%
Despite the wide exploration of various problems, improving reconstruction and synthesis quality via the frequency domain remains much less explored.

%\vspace{0.1cm}
\noindent
\textbf{Hard example processing.}
Hard example processing is widely explored in object detection and image classification to address the class imbalance problem.
A common solution is to use a bootstrapping technique called hard example mining~\cite{ohem,hempartmodel}, where a representative method is online hard example mining (OHEM)~\cite{ohem}. The training examples are sampled following the current loss of each example to modify the stochastic gradient descent. The model is encouraged to learn hard examples more to boost performance.
An alternative solution is focal loss~\cite{focalloss}, which is a scaled cross-entropy loss. The scaling factor down-weights the contribution of easy examples during training so that a model can focus on learning hard examples.
The proposed focal frequency loss is inspired by these techniques. %the proposed  is a natural extension to tackle hard frequencies that generative models lose most during training. In other words, instead of class imbalance, our loss is designed to handle frequency bias and dynamically focus the model on a sparse set of hard frequencies.

%%\vspace{0.1cm}
%\noindent
%\textbf{Generative models.}
%
%(AE, VAE, GAN, I2I, SIS, Briefly mention related ones)
%
%
%\vspace{0.1cm}
%\noindent
%\textbf{Frequency domain analysis of neural networks.}
%
%(Fourier Analysis (F-principle), Fourier features and spectrum bias, compression (DCT/DWT), other applications such as SR (frequency separation), Forensics (CNN detection, etc.), MRI (joint frequency))
%
%
%\vspace{0.1cm}
%\noindent
%\textbf{Hard example processing.}
%
%(OHEM (online hard example mining, a simple modification to SGD in which training examples are sampled according to a non-uniform, non-stationary distribution that depends on the current loss of each example under consideration), focal loss (re-weighting, scaled cross entropy loss, down-weight the contribution of easy examples during training), etc. Rephrase here.)
% !TEX root = ../main_arxiv.tex

\section{Focal Frequency Loss}
\label{sec:method}

\if 0
We propose a \textit{focal frequency loss} to improve \liming{image reconstruction and synthesis quality} by narrowing the gap between the real and fake images in the frequency domain.
%
The loss allows the model to dynamically locate and focus on hard frequencies, \ie, the frequency components that are hard to synthesize.
\fi
%
%we explicitly utilize the frequency representation of images (Section 3.1) to find the hard frequencies easily.
To formulate our method, we explicitly exploit the frequency representation of images (Section~\ref{sec:freqrepre}), facilitating the network to locate the hard frequencies.
We then define a frequency distance (Section~\ref{sec:freqdist}) to quantify the differences between images in the frequency domain.
Finally, we adopt a dynamic spectrum weighting scheme (Section~\ref{sec:weightmatrix}) that allows the model to focus on the on-the-fly hard frequencies.% during training.

\subsection{Frequency Representation of Images}
\label{sec:freqrepre}

%We explicitly use the frequency representation of images to help the generative models find the hard frequencies readily during training.
%
In this section, we revisit and highlight several key concepts of the discrete Fourier transform. We demonstrate the effect of missing frequencies in the image and the advantage of frequency representation for locating hard frequencies.
%
%In this section, we revisit and highlight several key concepts of the discrete Fourier transform and the physical meanings of different regions on the frequency spectrum.
%%
%We demonstrate the effect of missing frequencies in the image and the advantage of frequency representation for locating the hard frequencies.

Discrete Fourier transform (DFT) is a complex-valued function that converts a discrete finite signal into its constituent frequencies, \ie, complex exponential waves.
An image\footnote{\hspace{0.07cm}For simplicity, the formulas in this section are applied to gray-scale images, while the extension to color images is straightforward by processing each channel separately in the same way.} can be treated as a two-dimensional discrete finite signal with only real numbers. Thus, to convert an image into its frequency representation, we perform the 2D discrete Fourier transform:
\begin{equation}
\label{eq:1}
    F\left(u,v\right)=\sum_{x=0}^{M-1}{\sum_{y=0}^{N-1}{f\left(x,y\right)\cdot e^{-i2\pi\left(\frac{ux}{M}+\frac{vy}{N}\right)}}},
\end{equation}
where the image size is $M \times N$; $\left(x, y\right)$ denotes the coordinate of an image pixel in the spatial domain; $f\left(x, y\right)$ is the pixel value; $\left(u, v\right)$ represents the coordinate of a spatial frequency on the frequency spectrum;  $F\left(u,v\right)$ is the complex frequency value; $e$ and $i$ are Euler's number and the imaginary unit, respectively.
Following Euler's formula:
\begin{equation}
\label{eq:2}
    e^{i\theta}=\cos{\theta}+i\sin{\theta},
\end{equation}
the natural exponential function in Eq.~\eqref{eq:1} can be written as:
\begin{equation}
\label{eq:3}
\footnotesize
    e^{-i2\pi\left(\frac{ux}{M}+\frac{vy}{N}\right)}
    =\cos{2\pi\left(\frac{ux}{M}+\frac{vy}{N}\right)}
    -i\sin{2\pi\left(\frac{ux}{M}+\frac{vy}{N}\right)}.
\end{equation}

According to Eq.~\eqref{eq:1} and Eq.~\eqref{eq:3}, the image is decomposed into orthogonal sine and cosine functions, constituting the imaginary and the real part of the frequency value, respectively, after applied 2D DFT.
%
%\wayne{Why introduce Eq.2 and Eq.3 here? To help understand the motivation of frequency loss? Is it necessary to give the definition detail here?}
%
Each sine or cosine can be regarded as a binary function of $\left(x, y\right)$, where its angular frequency is decided by the spectrum position $\left(u, v\right)$.
The mixture of these sines and cosines provides both the horizontal and vertical frequencies of an image.
Therefore, spatial frequency manifests as the 2D sinusoidal components in the image.
The spectrum coordinate $\left(u, v\right)$ also represents the angled direction of a spatial frequency (visualizations can be found in the \textit{Appendix}), and $F\left(u, v\right)$ shows the ``response'' of the image to this frequency.
Due to the periodicity of trigonometric functions, the frequency representation of an image also acquires periodic properties.

\begin{figure}[t]
	\centering
%	\vspace{-0.35cm}
	\includegraphics[width=\linewidth]{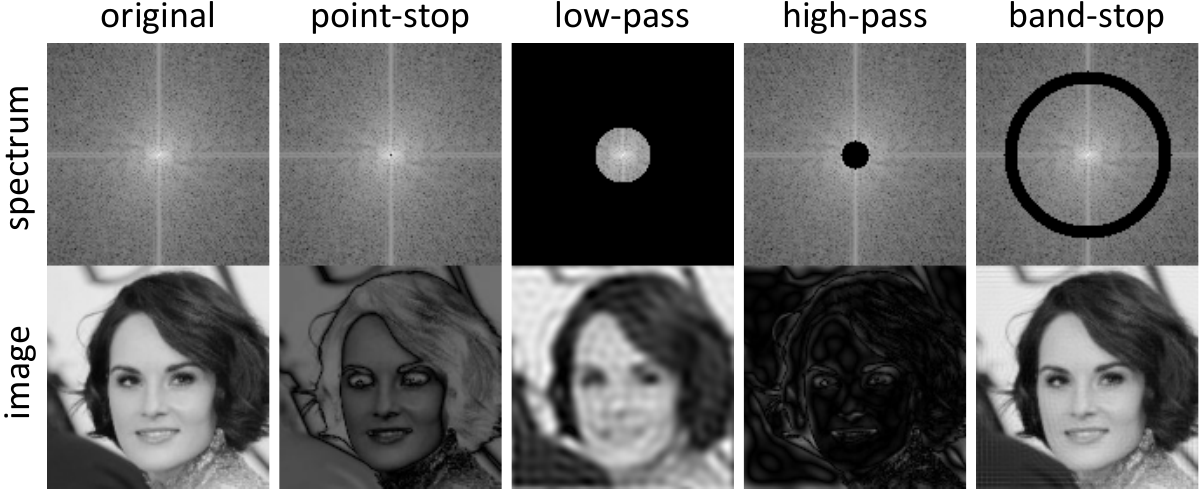}
%	\vspace{-0.5cm}
	\caption{Standard bandlimiting operations on the frequency spectrum with the origin (low frequencies) center shifted and respective images in the spatial domain. These manual operations can be regarded as a simulation to show the effect of missing frequencies.}
	\label{fig:spectrumstudy}
	\vspace{-0.35cm}
\end{figure}

Note that in Eq.~\eqref{eq:1}, $F\left(u,v\right)$ is the sum of a function that traverses every image pixel in the spatial domain, hence a specific spatial frequency on the spectrum depends on all the image pixels.
For an intuitive visualization, we suppress the \textit{single} center point (the lowest frequency) of the spectrum (Column $2$ of Figure~\ref{fig:spectrumstudy}), leading to \textit{all} the image pixels being affected.
To further ascertain the spatial frequency at the different regions on the spectrum, we perform some other standard bandlimiting operations and visualize their physical meanings in the spatial domain (Figure~\ref{fig:spectrumstudy}).
A low-pass filter (Column $3$), \ie, missing high frequencies, causes blur and typical ringing artifacts. A high-pass filter (Column $4$), \ie, missing low frequencies, tends to retain edges and boundaries. Interestingly, a simple band-stop filter (Column $5$), \ie, missing certain frequencies, produces visible common checkerboard artifacts (zoom in for view).%, which are quite common for generative models.
%
%\wayne{Empirical study can make it more intuitive. But, a theoretical rather than empirical analysis may need to motivate the add of frequency loss.}
%

Observably, the losses of different regions on the frequency spectrum correspond to different artifacts on the image.
One may deduce that compensating for these losses may reduce artifacts and improve image reconstruction and synthesis quality.
%
%Besides, explicitly utilizing the frequency representation of images can facilitate the hard frequency locating on the spectrum.
%
The analysis here shows the value of using the frequency representation of images for profiling and locating different frequencies, especially the hard ones.

%\begin{figure}[t]
%	\centering
%%	\vspace{-0.35cm}
%	\includegraphics[width=\linewidth]{figures/spatialfreq.pdf}
%%	\vspace{-0.5cm}
%	\caption{Two-dimensional sinusoidal components, \ie, spatial frequencies, in an image. The angled direction and density (angular frequency) of the waves depend on the spectrum coordinate $\left(u, v\right)$.}
%	\label{fig:spatialfreq}
%%	\vspace{-0.3cm}
%\end{figure}

\subsection{Frequency Distance}
\label{sec:freqdist}

To devise a loss function for the missing frequencies, we need a distance metric that quantifies the differences between real and fake images in the frequency domain. The distance has to be differentiable to support stochastic gradient descent.
In the frequency domain, the data objects are different spatial frequencies on the frequency spectrum, appearing as different 2D sinusoidal components in an image.
%
%As mentioned in Section \ref{sec:freqrepre}, the spectrum coordinate $\left(u, v\right)$ determines the angled direction and the angular frequency of the sinusoidal waves (see Figure \ref{fig:spatialfreq}).
%%
%Besides, the frequency value $F\left(u,v\right)$, a complex number, shows the ``reflection'' of the image to the spatial frequency at $\left(u, v\right)$.
%
To design our frequency distance, we further study the real and imaginary part of the complex value $F\left(u,v\right)$ in Eq.~\eqref{eq:1}.

Let $R\left(u,v\right)=a$ and $I\left(u,v\right)=b$ be the real and the imaginary part of $F\left(u,v\right)$, respectively. $F\left(u,v\right)$ can be rewritten as:
\begin{equation}
\label{eq:4}
    F\left(u,v\right)=R\left(u,v\right)+I\left(u,v\right)i=a+bi.
\end{equation}
According to the definition of 2D discrete Fourier transform, there are two key elements in $F\left(u,v\right)$.% that shows the image ``reflection'' to the certain spatial frequency at $\left(u,v\right)$.
The first element is \textit{amplitude}, which is defined as:
\begin{equation}
\label{eq:5}
    \left|F\left(u,v\right)\right|=\sqrt{R\left(u,v\right)^2+I\left(u,v\right)^2}=\sqrt{a^2+b^2}.
\end{equation}
Amplitude manifests the energy, \ie, how strongly an image responds to the 2D sinusoidal wave with a specific frequency. We typically show the amplitude as an informative visualization of the frequency spectrum (\eg, Figure~\ref{fig:teaser} and \ref{fig:spectrumstudy}).
The second element is \textit{phase}, which is written as:
\begin{equation}
\label{eq:6}
    \angle F\left(u,v\right)=\arctan{\left(\frac{I\left(u,v\right)}{R\left(u,v\right)}\right)}=\arctan{\frac{b}{a}}.
\end{equation}
Phase represents the shift of a 2D sinusoidal wave from the wave with the origin value (the beginning of a cycle).

\begin{figure}[t]
	\centering
%	\vspace{-0.35cm}
	\includegraphics[width=\linewidth]{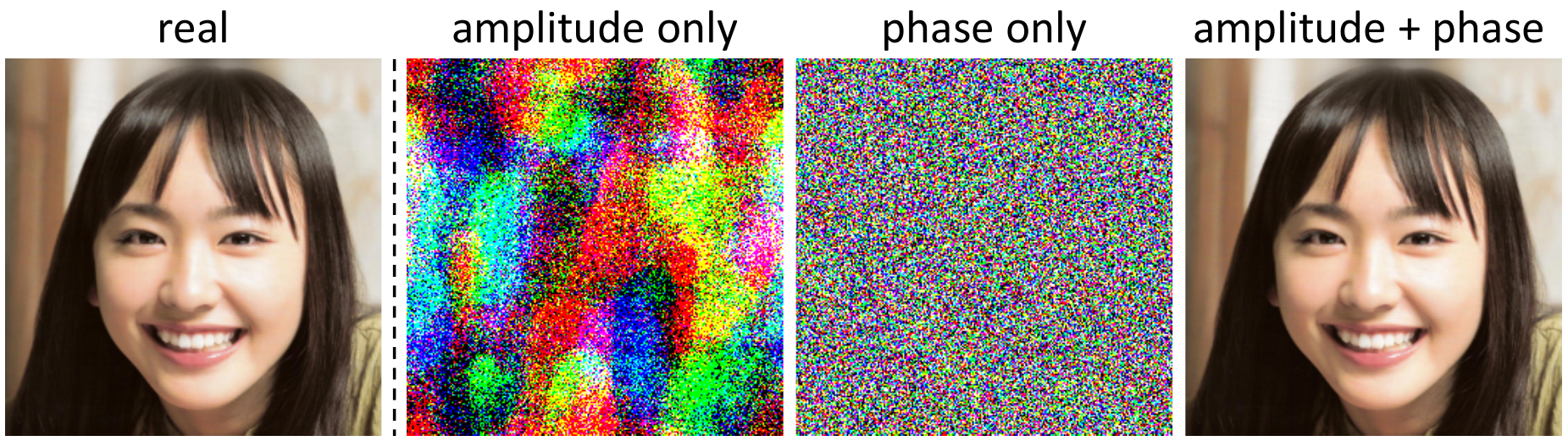}
%	\vspace{-0.5cm}
	\caption{The necessity of both amplitude and phase information for a frequency distance verified by \textit{single-image reconstruction}. ``Amplitude/phase only'' means solely applying Eq.~\eqref{eq:5}/\eqref{eq:6} to calculate the distance between the real and reconstructed images.}
	\label{fig:ampliphase}
	\vspace{-0.3cm}
\end{figure}

A frequency distance should consider both the amplitude and the phase as they capture different information of an image.
We show a single-image reconstruction experiment in Figure~\ref{fig:ampliphase}.
Merely minimizing the amplitude difference returns a reconstructed image with irregular color patterns.
Conversely, using only the phase information, the synthesized image resembles a noise.
A faithful reconstruction can only be achieved by considering both amplitude and phase.

\if 0
Since amplitude represents the energy of the image for a certain spatial frequency, which is suitable to visualize the spectrum difference, we initially consider using the amplitude information to design a frequency distance.
%
%At first, we start to consider using the amplitude information to design a frequency distance, since it represents the energy of the image for a certain spatial frequency, which is suitable for the spectrum visualization.
%
However, the amplitude only is not sufficient to retain the entire image information.
%
For verification, we perform a simple single-image reconstruction task by minimizing the amplitude difference (Column $2$ of Figure \ref{fig:ampliphase}).
%
The reconstructed image becomes irregular color patterns.
%
Thus, the phase is also essential to retain the image information.
%
Conversely, if we only use the phase information (Column $3$), the synthesized image becomes a noise.
%
Therefore, the frequency distance should consider both the amplitude and phase information.
%
%\wayne{We should have some explanation about why we use both amplitude and phase. Miss the information about why phase is also important.}
%
%\daibo{it's better if you can use some examples or an ablation study to support this point. more details should be included about why amplitude/phase is important. currently there are only definitions.}
\fi

%(A simple choice is to use x/y, but we hope to define a single-valued distance for each frequency which may be more helpful to quantifies the frequency loss.)

\begin{figure}[t]
	\centering
%	\vspace{-0.35cm}
	\includegraphics[width=0.9\linewidth]{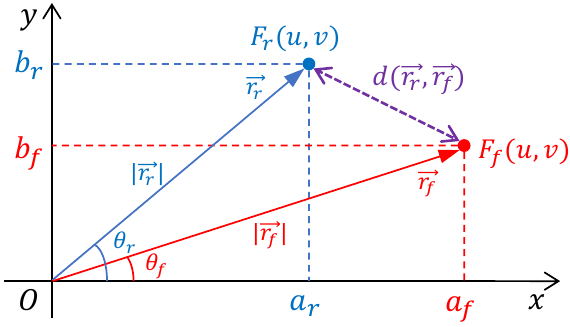}
%	\vspace{-0.5cm}
	\caption{Frequency distance between $\vec{r_r}$ and $\vec{r_f}$ mapped from two corresponding real and fake frequency values $F_r\left(u,v\right)$ and $F_f\left(u,v\right)$ at the spectrum position $\left(u,v\right)$. The Euclidean distance (purple line) is used, considering both the amplitude (magnitude $|\vec{r_r}|$ and $|\vec{r_f}|$) and phase (angle $\theta_r$ and $\theta_f$) information.}
	\label{fig:freqdist}
	\vspace{-0.2cm}
\end{figure}

Our solution is to map each frequency value to a Euclidean vector in a two-dimensional space (\ie, a plane). Following the standard definition of a complex number, the real and imaginary parts correspond to the $x$-axis and $y$-axis, respectively.
Let $F_r\left(u,v\right)=a_r+b_ri$ be the spatial frequency value at the spectrum coordinate $\left(u,v\right)$ of the real image, and the corresponding $F_f\left(u,v\right)=a_f+b_fi$ with the similar meaning \wrt the fake image.
We denote $\vec{r_r}$ and $\vec{r_f}$ as two respective vectors mapped from $F_r\left(u,v\right)$ and $F_f\left(u,v\right)$ (see Figure~\ref{fig:freqdist}).
Based on the definition of amplitude and phase, we note that the vector magnitude $|\vec{r_r}|$ and $|\vec{r_f}|$ correspond to the amplitude, and the angle $\theta_r$ and $\theta_f$ correspond to the phase.
Thus, the frequency distance corresponds to the distance between $\vec{r_r}$ and $\vec{r_f}$, which considers both the vector magnitude and angle.
We use the (squared) Euclidean distance for a single frequency:
\begin{equation}
\label{eq:7}
    d\left(\vec{r_r},\vec{r_f}\right)=\|\vec{r_r}-\vec{r_f}\|_2^2=|F_r\left(u,v\right)-F_f\left(u,v\right)|^2.
\end{equation}
The frequency distance between the real and fake images can be written as the average value: 
\begin{equation}
\small
\label{eq:8}
    d\left(F_r,F_f\right)=\frac{1}{MN}\sum_{u=0}^{M-1}\sum_{v=0}^{N-1}{|F_r\left(u,v\right)-F_f\left(u,v\right)|^2}.
\end{equation}
%

\if 0
This frequency distance quantifies the difference between the real and fake in the frequency domain, with both the amplitude and phase information under consideration, meanwhile supporting the stochastic gradient descent.
%
By minimizing this frequency distance, both the amplitude and the phase difference will be reduced. Therefore, the reconstructed image (Column $4$ of Figure \ref{fig:ampliphase}) will be very close to the original real image (Column $1$).
\fi

%\subsection{Spectrum Weight Matrix}
\subsection{Dynamic Spectrum Weighting}
\label{sec:weightmatrix}

%
%\wayne{We may need a figure to help understanding the design of focal frequency loss.}
%

\if 0
As discussed in Section \ref{sec:freqrepre}, by explicitly transforming images to their frequency representations using Eq.~\eqref{eq:1}, networks can easily find the hard frequencies during training.
%
In this section, we hope to tackle these hard frequencies for generative models to improve the synthesis quality.
\fi

The frequency distance we defined in Eq.~\eqref{eq:8} quantitatively compares the real and fake images in the frequency domain.
%Based on the frequency distance we defined in Eq.~\eqref{eq:8}, the model can quantitatively compare the real and fake images in the frequency domain instead of the spatial domain.
%
However, a direct use of Eq.~\eqref{eq:8} as a loss function is not helpful in coping with hard frequencies since the weight of each frequency is identical. A model would still bias to easy frequencies due to the inherent bias. % of the networks.

Inspired by hard example mining~\cite{hempartmodel,ohem} and focal loss~\cite{focalloss}, we formulate our method to focus the training on the hard frequencies. To implement this, we introduce a spectrum weight matrix to down-weight the easy frequencies.
The spectrum weight matrix is dynamically determined by a non-uniform distribution on the current loss of each frequency during training.
Each image has its own spectrum weight matrix.
The shape of the matrix is the same as that of the spectrum.
The matrix element $w\left(u,v\right)$, \ie, the weight for the spatial frequency at $\left(u,v\right)$, is defined as:
\begin{equation}
\label{eq:9}
    w\left(u,v\right)=|F_r\left(u,v\right)-F_f\left(u,v\right)|^\alpha,
\end{equation}
where $\alpha$ is the scaling factor for flexibility ($\alpha=1$ in our experiments). We further normalize the matrix values into the range $[0, 1]$, where the weight $1$ corresponds to the currently most lost frequency, and the easy frequencies are down-weighted. The gradient through the spectrum weight matrix is locked, so it only serves as the weight for each frequency.

By performing the Hadamard product for the spectrum weight matrix and the frequency distance matrix, we have the \textit{full form} of the focal frequency loss (FFL):
\begin{equation}
\small
\label{eq:10}
    \mathrm{FFL}=\frac{1}{MN}\sum_{u=0}^{M-1}\sum_{v=0}^{N-1}{w\left(u,v\right)|F_r\left(u,v\right)-F_f\left(u,v\right)|^2}.
\end{equation}
The focal frequency loss can be seen as a weighted average of the frequency distance between the real and fake images.
It focuses the model on synthesizing hard frequencies by down-weighting easy frequencies. % using the introduced spectrum weight matrix.
Besides, the focused region is updated on the fly to complement the immediate hard frequencies, thus progressively refining the generated images and being adaptable to different methods.

In practice, to apply the proposed focal frequency loss to a model, we first transform both the real and fake images into their frequency presentations using the 2D DFT.
We then perform the orthonormalization for each frequency value $F\left(u,v\right)$, \ie, dividing it by $\sqrt{MN}$, so that the 2D DFT is unitary to ensure a smooth gradient.
Finally, we employ Eq.~\eqref{eq:10} to calculate the focal frequency loss.
We note that the exact form of focal frequency loss is not crucial.
%
%Several variants can be considered for the flexibility.
%
%Some simple variants can be derived by adjusting the spectrum weight matrix parameter $\alpha$ in Eq.~\eqref{eq:9}.
%%
%The parameter $\alpha$ controls how close the values of the weight matrix are, \ie, how focused the model is.
%%
%%and the parameter $\beta$ serves as the base value (minimum) of the matrix.
%%
%In our experiments, we set $\alpha=1$.
%%
%Besides, one can attempt patch-based focal frequency loss by cropping an image into small patches so that the focused frequencies are at the patch level.
%
Some studies on the loss variants are provided in the \textit{Appendix}.

\section{Experiments}
\label{sec:experiments}

\subsection{Settings}
\label{sec:settings}
%
%\vspace{0.1cm}
\noindent
\textbf{Baselines.}
%
%Our experiments cover different image reconstruction and synthesis methods.
%
We start from image reconstruction by vanilla AE~\cite{ae} (\ie, a simple 2-layer MLP) and VAE~\cite{vae} (\ie, CNN-based).
%We start from vanilla AE~\cite{ae} (\ie, a 2-layer MLP) to evaluate the ability of the focal frequency loss (FFL) to help image reconstruction on simple baselines.
%
We then study unconditional image synthesis using VAE, \ie, generating images from the Gaussian noise.
%
%We then vary the network structure to CNN using VAE~\cite{vae} for image reconstruction and unconditional image synthesis (\ie, generating images from the Gaussian noise).
%
Besides, we also investigate conditional image synthesis using GAN-based methods. Specifically, we select two typical image-to-image translation approaches, \ie, pix2pix~\cite{pix2pix} and SPADE~\cite{SPADE}.
% (\ie, generating images based on certain conditions)
We further explore the potential of focal frequency loss (FFL) on state-of-the-art StyleGAN2~\cite{stylegan2}.
In addition, we compare FFL with relevant losses~\cite{perceptualloss,specreg}.
%
%These baselines vary in categories, network structures, tasks, and spatial losses, to verify the effectiveness and versatility of FFL for various settings.
%
The implementation details are provided in the \textit{Appendix}.

\vspace{0.05cm}
\noindent
\textbf{Datasets.}
We use a total of seven datasets.
The datasets vary in types, sizes, and resolutions.
For vanilla AE, we exploit the Describable Textures Dataset (DTD)~\cite{DTD} and CelebA~\cite{celeba}.
%
%For vanilla AE, we exploit the Describable Textures Dataset (DTD)~\cite{DTD}, containing texture images with special frequency patterns. Besides, we use the cropped and aligned faces in CelebA~\cite{celeba}, which are more natural images.
%
For VAE, we use CelebA and CelebA-HQ~\cite{pggan} with different resolutions.
%
%For VAE, we exploit the cropped and aligned face images in CelebA and CelebA-HQ~\cite{pggan}. We preprocess the face images into different resolutions.
%
For pix2pix, we utilize the officially prepared CMP Facades~\cite{cmpfacades} and edges $\rightarrow$ shoes~\cite{shoesutzappos50K} datasets.
For SPADE, we select two challenging datasets, \ie, Cityscapes~\cite{cityscapes} and ADE20K~\cite{ade20k}.
For StyleGAN2, we reuse CelebA-HQ.
Please refer to the \textit{Appendix} for the dataset details.
%
%We conduct experiments on a total of seven datasets under diverse settings. The datasets vary in types, sizes, and resolutions.
%%
%For vanilla AE, we use the Describable Textures Dataset (DTD) \cite{DTD}, containing texture images with special frequency patterns. Besides, we use the cropped and aligned faces in CelebA \cite{celeba}, which are more natural images.
%%
%For VAE, we exploit the cropped and aligned face images in CelebA and CelebA-HQ \cite{pggan}. We preprocess the face images into different resolutions.
%%
%For pix2pix, we utilize the officially prepared CMP Facades \cite{cmpfacades} and edges $\rightarrow$ shoes \cite{shoesutzappos50K} datasets. The edge maps are detected by HED \cite{hededgedet}.
%%
%For SPADE, we select two challenging datasets, \ie, Cityscapes \cite{cityscapes} and ADE20K \cite{ade20k}.
%%
%Please refer to the \textit{supplementary material} for the dataset details.
%, including the source, number of images, preprocessing, resolution, \etc.

%(DTD, CelebA, CelebA-HQ, facades, edges2shoes, Cityscapes, ade20k.)

%\subsection{Evaluation Metrics}
%\label{sec:metrics}

\vspace{0.05cm}
\noindent
\textbf{Evaluation metrics.}
To evaluate frequency domain difference, we introduce a frequency-level metric, named Log Frequency Distance (LFD), which is defined by a modified version of Eq.~\eqref{eq:8}:
\begin{equation}
\footnotesize
\label{eq:11}
    \mathrm{LFD}=\log{\left[\frac{1}{MN}\left(\sum_{u=0}^{M-1}\sum_{v=0}^{N-1}{|F_r\left(u,v\right)-F_f\left(u,v\right)|^2}\right)+1\right]},
\end{equation}
where the logarithm is only used to scale the value into a reasonable range.
%
%LFD compares the real and fake images in the frequency domain.
%
A lower LFD is better. Note that LFD is a full reference metric (\ie, requiring the ground truth image), so we use it in the reconstruction tasks.

Besides, we integrate the evaluation protocols from prior works~\cite{nerf,BigGAN,SPADE,tsit}.
Specifically, we employ FID (lower is better)~\cite{TTUR} for all tasks.
For the reconstruction tasks of vanilla AE and VAE, we use PSNR (higher is better), SSIM (higher is better)~\cite{ssim}, and LPIPS (lower is better)~\cite{lpips} in addition to LFD and FID.
For the synthesis tasks of VAE, pix2pix, and StyleGAN2, we apply IS (higher is better)~\cite{is} in addition to FID.
For SPADE (task-specific method for semantic image synthesis), besides FID, we follow their paper~\cite{SPADE} to use mIoU (higher is better) and pixel accuracy (accu, higher is better) for the segmentation performance of synthesized images. We use DRN-D-105~\cite{drn} for Cityscapes and UperNet101~\cite{upernet} for ADE20K.

%(1) AE and VAE reconstruction: PSNR, SSIM, LPIPS, FID, LFD.
%
%(2) Pix2pix and VAE generation: FID, IS.
%
%(3) SPADE: mIoU, Accuracy, FID.

\subsection{Results and Analysis}
\label{sec:results}

\begin{figure}[t]
	\centering
%	\vspace{-0.35cm}
	\includegraphics[width=\linewidth]{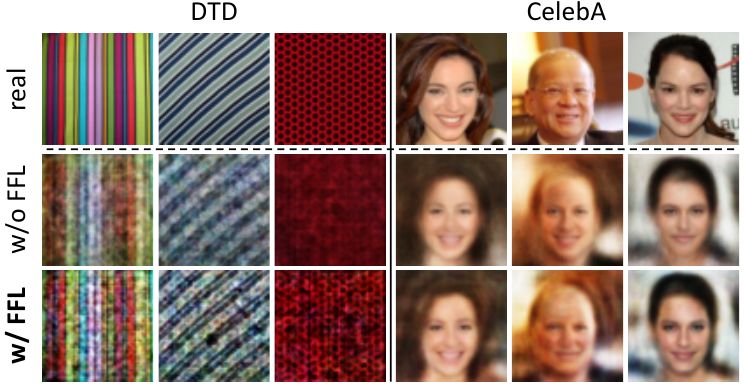}
%	\vspace{-0.5cm}
	\caption{\textbf{Vanilla AE image reconstruction} results on the \textbf{DTD} ($64 \times 64$) and \textbf{CelebA} ($64 \times 64$) datasets.}
	\label{fig:aerecon}
	\vspace{-0.1cm}
\end{figure}

%\begin{figure}[t]
%	\centering
%%	\vspace{-0.35cm}
%	\includegraphics[width=\linewidth]{figures/aedtd.pdf}
%%	\vspace{-0.5cm}
%	\caption{\textbf{Vanilla AE image reconstruction} results on the \textbf{DTD} ($64 \times 64$) dataset.}
%	\label{fig:aedtd}
%	\vspace{-0.2cm}
%\end{figure}
%
%
%\begin{figure}[t]
%	\centering
%%	\vspace{-0.35cm}
%	\includegraphics[width=\linewidth]{figures/aeceleba.pdf}
%%	\vspace{-0.5cm}
%	\caption{\textbf{Vanilla AE image reconstruction} results on the \textbf{CelebA} ($64 \times 64$) dataset.}
%	\label{fig:aeceleba}
%	\vspace{-0.1cm}
%\end{figure}

\begin{table}[tb!]
%\vspace{-0.1cm}
%\addtolength{\tabcolsep}{3pt}
\centering
\footnotesize
\caption{The PSNR (higher is better), SSIM (higher is better), LPIPS (lower is better), FID (lower is better) and LFD (lower is better) scores for the \textbf{vanilla AE image reconstruction} trained with/without the focal frequency loss (FFL).}
%\vspace{0.1cm}
\begin{tabularx}{\linewidth}{c|c|*{5}{|Y}}
\Xhline{1pt}
Dataset& FFL & PSNR$\uparrow$& SSIM$\uparrow$& LPIPS$\downarrow$& FID$\downarrow$& LFD$\downarrow$ \\
\cline{2-7}
%& \multicolumn{5}{c}{DTD} \\
\Xhline{0.6pt}
DTD& w/o & 20.133& {\bf0.407}& 0.414& 246.870& 14.764 \\
& w/ &  {\bf20.151}& 0.400& {\bf0.404}& {\bf240.373}& {\bf14.760} \\
\Xhline{0.6pt}
CelebA& w/o & 20.044& 0.568& 0.237& 97.035& 14.785 \\
& w/ &  {\bf21.703}& {\bf0.642}& {\bf0.199}& {\bf83.801}& {\bf14.403} \\
\Xhline{1pt}
\end{tabularx}
\label{tbl:vanillaae}
\vspace{-0.55cm}
\end{table}

%\vspace{0.1cm}
\noindent
\textbf{Vanilla AE.}
The results of vanilla AE~\cite{ae} image reconstruction are shown in Figure~\ref{fig:aerecon}.
On DTD, without the focal frequency loss (FFL), the vanilla AE baseline synthesizes blurry images, which lack sufficient texture details and only contain some low-frequency information.
With FFL, the reconstructed images become clearer and show more texture details.
The results on CelebA show that FFL improves a series of quality problems, \eg, face blur (Column $4$), identity shift (Column $5$), and expression loss (Column $6$).

The quantitative evaluation results are presented in Table~\ref{tbl:vanillaae}.
Adding the proposed FFL to the vanilla AE baseline leads to a performance boost in most cases on the DTD and CelebA datasets \wrt five evaluation metrics.
We note that the performance boost on CelebA is larger, indicating the effectiveness of FFL for the natural images.

%Using only a simple network structure (\ie, 2-layer MLP), the vanilla AE image reconstruction results are obviously improved by the focal frequency loss qualitatively and quantitatively, especially for the natural images.

%(qualitative examples, quantitative, training loss, freq gap, \etc)

\begin{figure}[t]
	\centering
	\vspace{-0.2cm}
	\includegraphics[width=\linewidth]{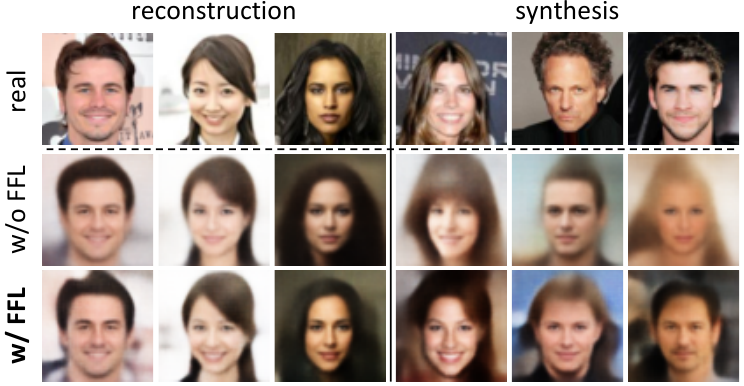}
%	\vspace{-0.1cm}
	\caption{\textbf{VAE image reconstruction} and \textbf{unconditional image synthesis} results on the \textbf{CelebA} ($64 \times 64$) dataset.}
	\label{fig:vaeceleba}
	\vspace{-0.35cm}
\end{figure}

\begin{figure}[t]
	\centering
%	\vspace{-0.35cm}
	\includegraphics[width=0.96\linewidth]{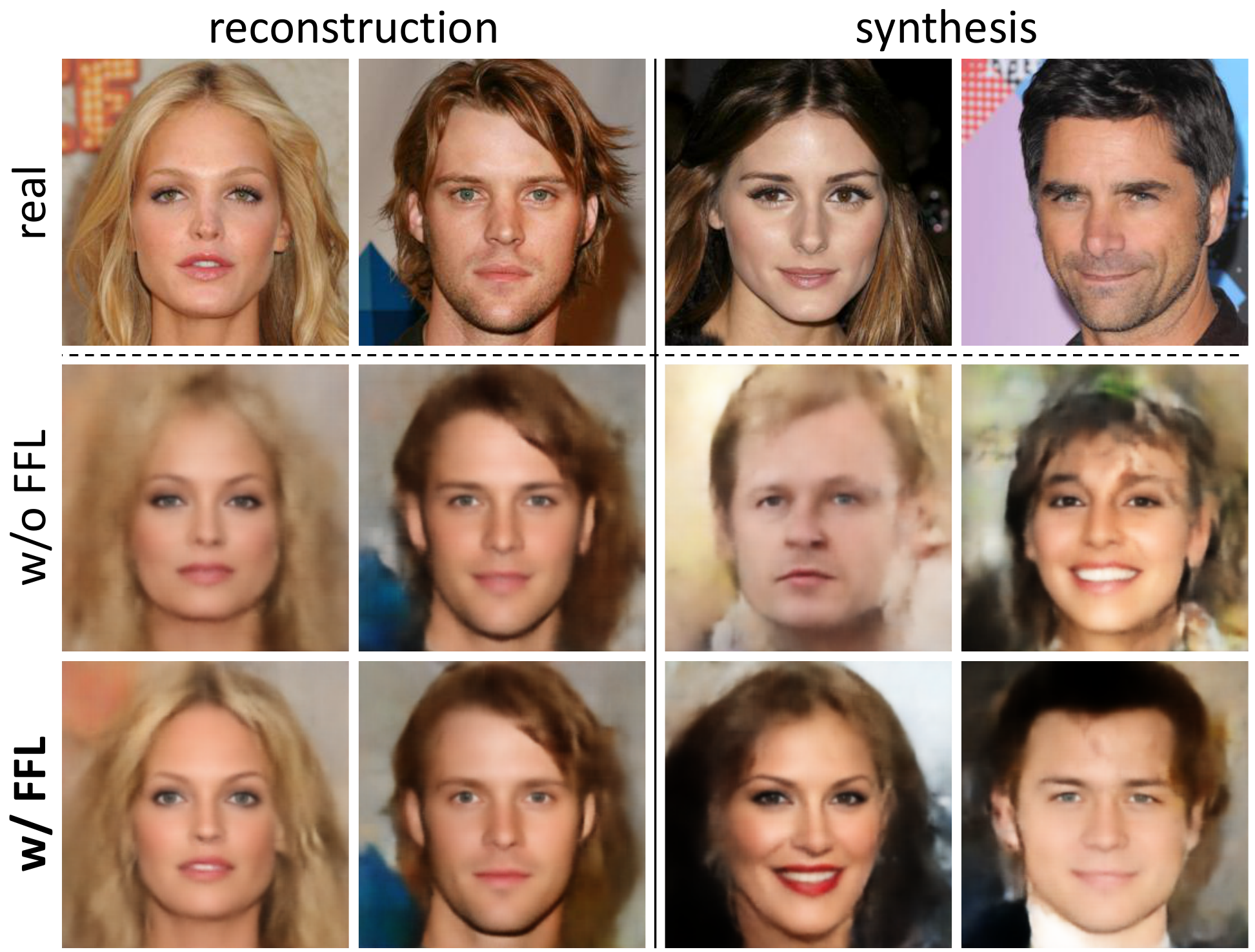}
	\vspace{-0.05cm}
	\caption{\textbf{VAE image reconstruction} and \textbf{unconditional image synthesis} results on the \textbf{CelebA-HQ} ($256 \times 256$) dataset.}
	\label{fig:vaecelebahq}
	\vspace{-0.12cm}
\end{figure}

\vspace{0.05cm}
\noindent
\textbf{VAE.}
The results of VAE~\cite{vae} image reconstruction and unconditional image synthesis on CelebA are shown in Figure~\ref{fig:vaeceleba}.
For reconstruction, FFL helps the VAE model better retain the image clarity (Column $1$), expression (Column~$2$), and skin color (Column $3$).
The unconditional synthesis results (Column $4$, $5$, $6$) show that the quality of generated images is improved after applying FFL. The generated faces become clearer and gain more texture details.
For a higher resolution, we present the VAE reconstruction and synthesis results on CelebA-HQ in Figure~\ref{fig:vaecelebahq}.
By adding FFL to the VAE baseline, the reconstructed images keep more original image information, \eg, mouth color (Column $2$) and opening angle (Column $1$). Besides, high-frequency details on the hair are clearly enhanced (Column $1$).
For unconditional image synthesis, FFL helps reduce artifacts and ameliorates the perceptual quality of synthesized images.

%\begin{figure}[t]
%	\centering
%%	\vspace{-0.35cm}
%	\includegraphics[width=\linewidth]{figures/vaecelebarec.pdf}
%%	\vspace{-0.1cm}
%	\caption{\textbf{VAE image reconstruction} results on the \textbf{CelebA} ($64 \times 64$) dataset.}
%	\label{fig:vaecelebarec}
%	\vspace{-0.2cm}
%\end{figure}
%
%
%\begin{figure}[t]
%	\centering
%%	\vspace{-0.35cm}
%	\includegraphics[width=\linewidth]{figures/vaecelebagen.pdf}
%%	\vspace{-0.5cm}
%	\caption{\textbf{VAE unconditional image synthesis} results on the \textbf{CelebA} ($64 \times 64$) dataset.}
%	\label{fig:vaecelebagen}
%	\vspace{-0.15cm}
%\end{figure}

\begin{table}[tb!]
%\vspace{-0.1cm}
%\addtolength{\tabcolsep}{3pt}
\centering
\footnotesize
\caption{The PSNR (higher is better), SSIM (higher is better), LPIPS (lower is better), FID (lower is better) and LFD (lower is better) scores for the \textbf{VAE image reconstruction} trained with/without the focal frequency loss (FFL).}
%\vspace{0.1cm}
\begin{tabularx}{\linewidth}{c|c|*{5}{|Y}}
\Xhline{1pt}
Dataset& FFL & PSNR$\uparrow$& SSIM$\uparrow$& LPIPS$\downarrow$& FID$\downarrow$& LFD$\downarrow$ \\
\cline{2-7}
%& \multicolumn{5}{c}{DTD} \\
\Xhline{0.6pt}
CelebA& w/o & 19.961& 0.606& 0.217& 69.900& 14.804 \\
& w/ &  {\bf22.954}& {\bf0.723}& {\bf0.143}& {\bf49.689}& {\bf14.115} \\
\Xhline{0.6pt}
CelebA-& w/o & 21.310& 0.616& 0.367& 71.081& 17.266 \\
HQ& w/ &  {\bf22.253}& {\bf0.637}& {\bf0.344}& {\bf59.470}& {\bf17.049} \\
\Xhline{1pt}
\end{tabularx}
\label{tbl:vaerec}
\vspace{-0.15cm}
\end{table}

%\begin{figure}[t]
%	\centering
%%	\vspace{-0.35cm}
%	\includegraphics[width=0.97\linewidth]{figures/vaecelebahqrec.pdf}
%	\vspace{-0.05cm}
%	\caption{\textbf{VAE image reconstruction} results on the \textbf{CelebA-HQ} ($256 \times 256$) dataset.}
%	\label{fig:vaecelebahqrec}
%	\vspace{-0.4cm}
%\end{figure}
%
%
%\begin{figure}[t]
%	\centering
%%	\vspace{-0.35cm}
%	\includegraphics[width=0.97\linewidth]{figures/vaecelebahqgen.pdf}
%%	\vspace{-0.5cm}
%	\caption{\textbf{VAE unconditional image synthesis} results on the \textbf{CelebA-HQ} ($256 \times 256$) dataset.}
%	\label{fig:vaecelebahqgen}
%	\vspace{-0.15cm}
%\end{figure}

\begin{table}[tb!]
%\vspace{-0.1cm}
%\addtolength{\tabcolsep}{3pt}
\centering
\footnotesize
\caption{The FID (lower is better) and IS (higher is better) scores for the \textbf{VAE unconditional image synthesis} trained with/without the focal frequency loss (FFL).}
%\vspace{0.1cm}
\begin{tabularx}{\linewidth}{c|c|*{2}{|Y}}
\Xhline{1pt}
Dataset& FFL & FID$\downarrow$& IS$\uparrow$ \\
\cline{2-4}
%& \multicolumn{5}{c}{DTD} \\
\Xhline{0.6pt}
CelebA& w/o & 80.116& 1.873 \\
& w/ &  {\bf71.050}& {\bf2.010} \\
\Xhline{0.6pt}
CelebA-& w/o & 93.778& 2.057 \\
HQ& w/ &  {\bf84.472}& {\bf2.060} \\
\Xhline{1pt}
\end{tabularx}
\label{tbl:vaegen}
\vspace{-0.6cm}
\end{table}

The quantitative test results of VAE image reconstruction are shown in Table~\ref{tbl:vaerec}. Adding FFL to the VAE baseline achieves better performance \wrt all the metrics.
Besides, both FID and IS are better in the unconditional image synthesis task (Table~\ref{tbl:vaegen}), indicating that the generated images are clearer and more photorealistic. The results suggests the effectiveness of the focal frequency loss in helping VAE to improve image reconstruction and synthesis quality.

\begin{figure}[t]
	\centering
	\vspace{-0.2cm}
	\includegraphics[width=0.97\linewidth]{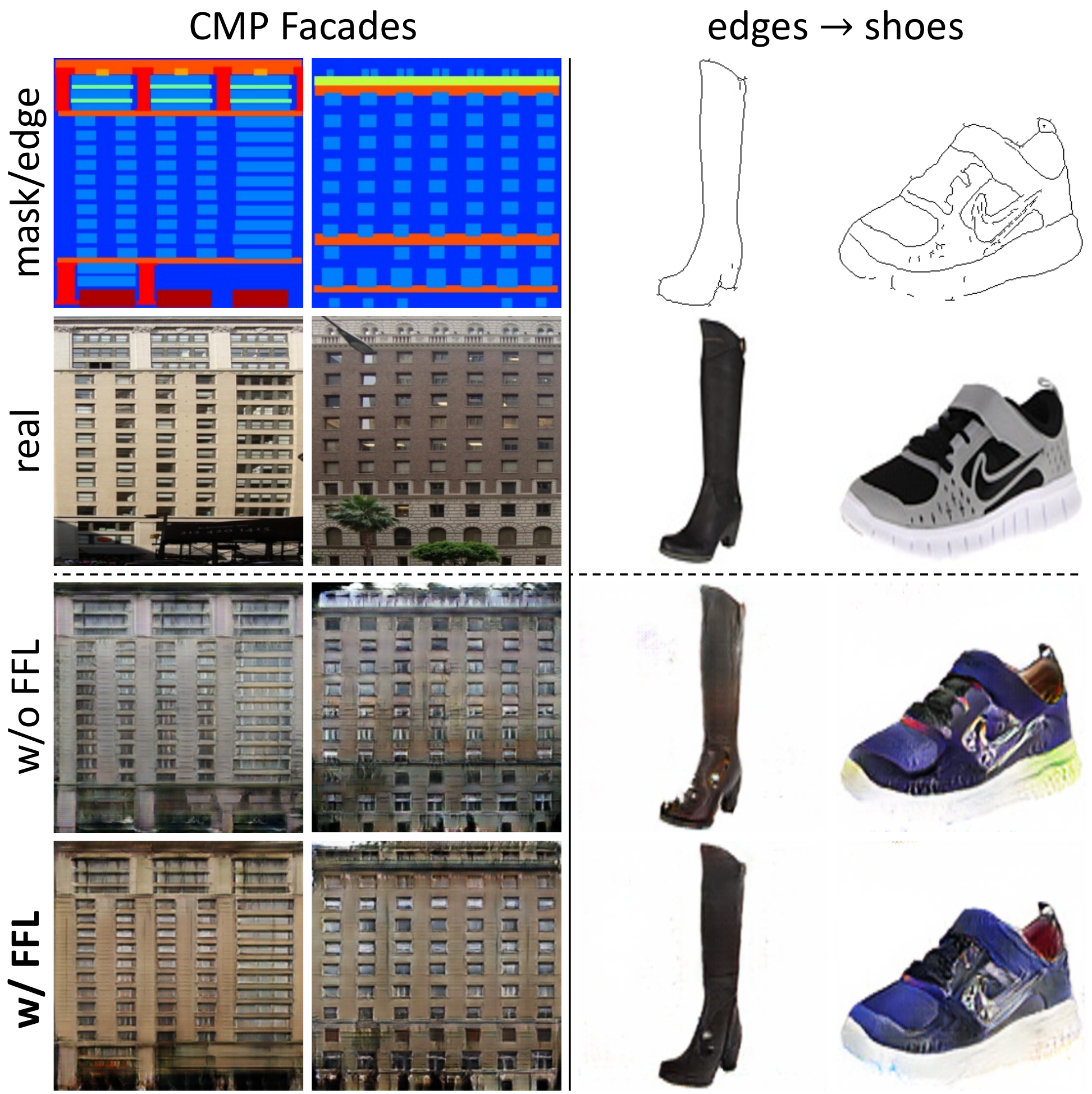}
%	\vspace{-0.05cm}
	\caption{\textbf{pix2pix image-to-image translation} results on \textbf{CMP Facades} ($256 \times 256$) and \textbf{edges $\rightarrow$ shoes} ($256 \times 256$) datasets.}
	\label{fig:pix2pix}
	\vspace{-0.2cm}
\end{figure}

\vspace{0.05cm}
\noindent
\textbf{pix2pix.}
For conditional image synthesis, the results of pix2pix~\cite{pix2pix} image-to-image translation (I2I) are shown in Figure~\ref{fig:pix2pix}.
On CMP Facades, FFL improves the image synthesis quality of pix2pix by reducing unnatural colors (Column $1$) or the black artifacts on the building (Column $2$). Meanwhile, the semantic information alignment with the mask becomes better after applying FFL.
For the edges $\rightarrow$ shoes translation, pix2pix baseline sometimes introduces colored checkerboard artifacts to the white background (Column $3$, zoom in for view). Besides, atypical colors appear in certain cases (Column $4$). In comparison, the model trained with FFL yields fewer artifacts.

The quantitative evaluation results of pix2pix image-to-image translation are shown in Table~\ref{tbl:pix2pix}. FFL contributes to a performance boost on both of the two datasets. The results of the pix2pix baseline show the adaptability of the focal frequency loss for the image-to-image translation problem.

\begin{table}[tb!]
%\vspace{-0.1cm}
%\addtolength{\tabcolsep}{3pt}
\centering
\footnotesize
\caption{The FID (lower is better) and IS (higher is better) scores for the \textbf{pix2pix image-to-image translation} trained with/without the focal frequency loss (FFL).}
%\vspace{0.1cm}
\begin{tabularx}{\linewidth}{c|c|*{2}{|Y}}
\Xhline{1pt}
Dataset& FFL & FID$\downarrow$& IS$\uparrow$ \\
\cline{2-4}
%& \multicolumn{5}{c}{DTD} \\
\Xhline{0.6pt}
CMP Facades& w/o & 128.492& 1.571 \\
& w/ &  {\bf123.773}& {\bf1.738} \\
\Xhline{0.6pt}
edges $\rightarrow$ shoes& w/o & 80.279& 2.674 \\
& w/ &  {\bf74.359}& {\bf2.804} \\
\Xhline{1pt}
\end{tabularx}
\label{tbl:pix2pix}
\vspace{-0.6cm}
\end{table}

\begin{figure*}[t]
	\centering
	\vspace{-0.2cm}
	\includegraphics[width=0.98\linewidth]{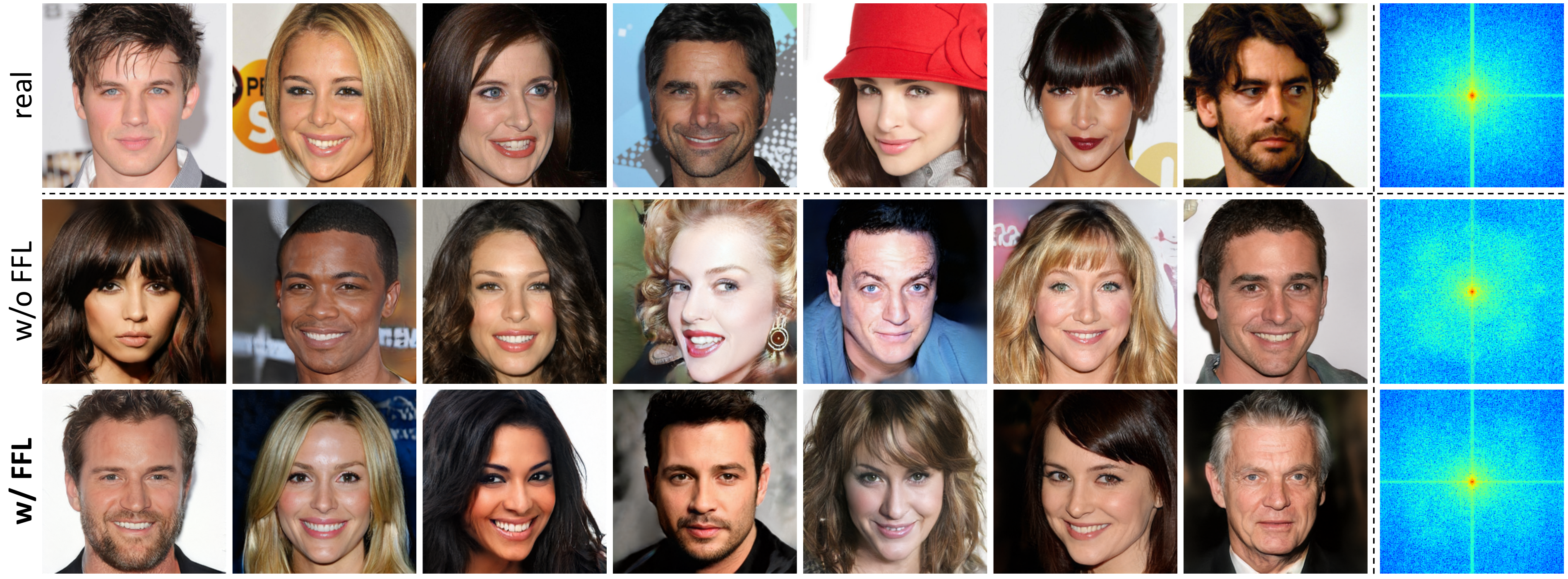}
	\vspace{-0.05cm}
	\caption{\textbf{StyleGAN2 unconditional image synthesis} results (without truncation) and the mini-batch average spectra (adjusted to better contrast) on the \textbf{CelebA-HQ} ($256 \times 256$) dataset.}
	\label{fig:stylegan2celebahq}
	\vspace{-0.4cm}
\end{figure*}

\begin{figure}[t]
	\centering
	\vspace{-0.1cm}
	\includegraphics[width=0.99\linewidth]{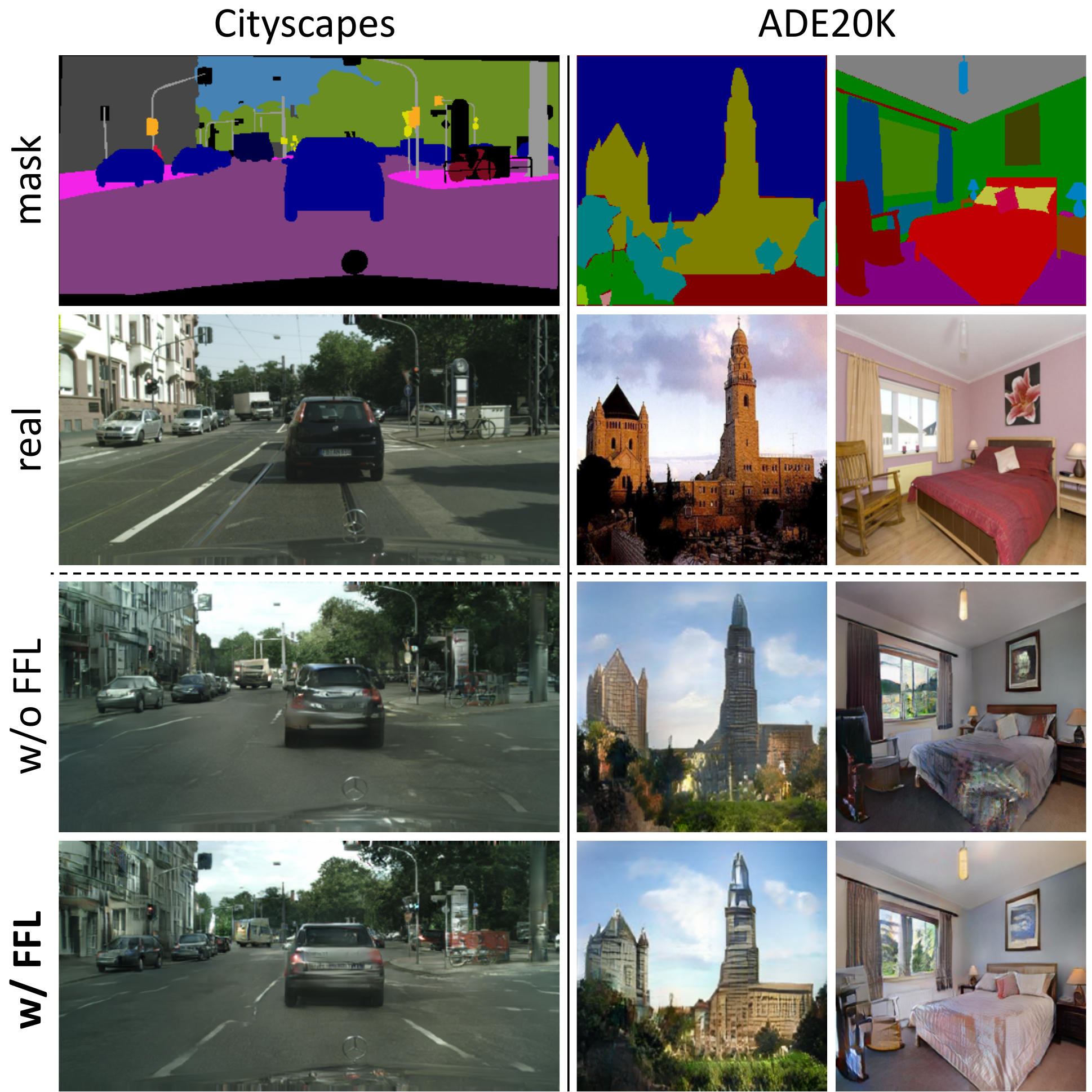}
%	\vspace{-0.5cm}
	\caption{\textbf{SPADE semantic image synthesis} results on the \textbf{Cityscapes} ($512 \times 256$) and \textbf{ADE20K} ($256 \times 256$) datasets.}
	\label{fig:spade}
	\vspace{-0.46cm}
\end{figure}

\vspace{0.05cm}
\noindent
\textbf{SPADE.}
We further explore semantic image synthesis (\ie, synthesizing a photorealistic image from a semantic segmentation mask) on more challenging datasets.
The results of SPADE~\cite{SPADE} are shown in Figure~\ref{fig:spade}.
In the street scene of Cityscapes (Column $1$), SPADE baseline distorts the car and road, missing some important details (\eg, road line).
The model trained with FFL demonstrates better perceptual quality for these details.
In the outdoor scene of ADE20K (Column $2$), applying FFL to SPADE boosts its ability to generate details on the buildings.
Besides, for the ADE20K indoor images (Column $3$), SPADE baseline produces some abnormal artifacts in certain cases. The model trained with the proposed FFL synthesizes more photorealistic images.

The quantitative test results are presented in Table~\ref{tbl:spade} (the values used for comparison are taken from~\cite{SPADE}).
We compare SPADE trained with/without FFL against a series of open-source task-specific semantic image synthesis methods~\cite{CRN,SIMS,pix2pixhd}.
SIMS~\cite{SIMS} obtains the best FID but poor segmentation scores on Cityscapes in that it directly stitches the training image patches from a memory bank while not keeping the exactly consistent positions.
Without modifying the SPADE network structure, training with FFL contributes a further performance boost, greatly outperforming the benchmark methods, which suggests the effectiveness of FFL for semantic image synthesis.

\begin{table}[tb!]
\vspace{-0.05cm}
%\addtolength{\tabcolsep}{3pt}
\centering
\footnotesize
\caption{The mIoU (higher is better), pixel accuracy (accu, higher is better) and FID (lower is better) scores for the \textbf{SPADE semantic image synthesis} trained with/without the focal frequency loss (FFL) compared to a series of task-specific methods.}
\begin{tabularx}{\linewidth}{l|*{6}{|Y}}
\Xhline{1pt}
& \multicolumn{3}{c}{Cityscapes} & \multicolumn{3}{|c}{ADE20K} \\
\cline{2-7}
Method& mIoU$\uparrow$& accu$\uparrow$& FID$\downarrow$& mIoU$\uparrow$& accu$\uparrow$& FID$\downarrow$ \\
\Xhline{0.6pt}
CRN~\cite{CRN} & 52.4& 77.1& 104.7& 22.4& 68.8& 73.3 \\
SIMS~\cite{SIMS} & 47.2& 75.5& {\bf49.7} & N/A& N/A& N/A \\
pix2pixHD~\cite{pix2pixhd} & 58.3& 81.4& 95.0& 20.3& 69.2& 81.8 \\
\Xhline{0.4pt}
SPADE~\cite{SPADE} &  62.3& 81.9& 71.8& 38.5& 79.9& 33.9 \\
SPADE + FFL &  {\bf64.2}& {\bf82.5}& \underline{59.5}& {\bf42.9}& {\bf82.4}& {\bf33.7} \\
\Xhline{1pt}
\end{tabularx}
\label{tbl:spade}
\vspace{-0.15cm}
\end{table}

\begin{table}[tb!]
%\vspace{-0.1cm}
%\addtolength{\tabcolsep}{3pt}
\centering
\footnotesize
\caption{The FID (lower is better) and IS (higher is better) scores for the \textbf{StyleGAN2 unconditional image synthesis} trained with/without the focal frequency loss (FFL).}
%\vspace{0.1cm}
\begin{tabularx}{\linewidth}{c|c|*{2}{|Y}}
\Xhline{1pt}
Dataset& FFL & FID$\downarrow$& IS$\uparrow$ \\
\cline{2-4}
%& \multicolumn{5}{c}{DTD} \\
\Xhline{0.6pt}
CelebA-HQ& w/o & 5.696& 3.383 \\
($256 \times 256$)& w/ & {\bf4.972}& {\bf3.432} \\
\Xhline{1pt}
\end{tabularx}
\label{tbl:stylegan2celebahq}
\vspace{-0.5cm}
\end{table}

\begin{table*}[tb!]
\centering
\footnotesize
\vspace{-0.15cm}
\caption{\textbf{Comparison} of our focal frequency loss (FFL) \textbf{with relevant losses}, \ie, perceptual loss (PL), spectral regularization (SpReg), and another transformation form for FFL, \ie, discrete cosine transform (DCT), in different image reconstruction and synthesis tasks.}
%\vspace{-0.35cm}
\begin{tabularx}{\linewidth}{l|*{9}{|Y}}
\Xhline{1pt}
& \multicolumn{5}{c}{VAE reconstruction (CelebA)}& \multicolumn{2}{|c}{VAE synthesis (CelebA)}& \multicolumn{2}{|c}{pix2pix I2I (edges $\rightarrow$ shoes)} \\
\cline{2-10}
Method& PSNR$\uparrow$& SSIM$\uparrow$& LPIPS$\downarrow$& FID$\downarrow$& LFD$\downarrow$& FID$\downarrow$& IS$\uparrow$& FID$\downarrow$& IS$\uparrow$ \\
\Xhline{0.6pt}
baseline & 19.961& 0.606& 0.217& 69.900& 14.804& 80.116& 1.873& 80.279& 2.674 \\
+ PL~\cite{perceptualloss} & 20.964& 0.658& {\bf0.143}& 62.795& 14.573& 78.825& 1.788& 78.916& 2.722 \\
+ SpReg~\cite{specreg} & 19.974& 0.607& 0.218& 69.118& 14.796& 78.079& 1.898& 79.300& 2.700 \\
\Xhline{0.4pt}
+ FFL (DCT) & 22.677& 0.711& 0.150& 51.536& 14.179& 71.827& 1.932& 79.045& 2.754 \\
+ FFL (Ours) & {\bf22.954}& {\bf0.723}& {\bf0.143}& {\bf49.689}& {\bf14.115}& {\bf71.050}& {\bf2.010}& {\bf74.359}& {\bf2.804} \\
\Xhline{1pt}
\end{tabularx}
\label{tbl:losscompare}
\vspace{-0.45cm}
\end{table*}

\vspace{0.05cm}
\noindent
\textbf{StyleGAN2.}
We apply FFL to the mini-batch average spectra of the real images and the generated images by the state-of-the-art unconditional image synthesis method, \ie, StyleGAN2~\cite{stylegan2}, intending to narrow the frequency distribution gap and improve quality further.
The results on CelebA-HQ ($256 \times 256$) without truncation~\cite{stylegan,stylegan2} are shown in Figure~\ref{fig:stylegan2celebahq}. Although StyleGAN2 (w/o FFL) generates photorealistic images in most cases, some tiny artifacts can still be spotted on the background (Column $2$ and $4$) and face (Column $5$).
Applying FFL, such artifacts are reduced, ameliorating synthesis quality further. Observably, the frequency domain gaps between mini-batch average spectra are clearly mitigated by FFL (Column $8$).
Some higher-resolution results can be found in the \textit{Appendix}.

The quantitative results are reported in Table~\ref{tbl:stylegan2celebahq}. FFL improves both FID and IS, in line with the visual quality enhancement. The results on StyleGAN2 show the potential of FFL to boost state-of-the-art baseline performance.

%(how to apply: mini-batch average spectrum, results, freq disc, higher res)

\vspace{0.05cm}
\noindent
\textbf{Comparison with relevant losses.}
For completeness and fairness, we compare the proposed focal frequency loss (FFL) with relevant loss functions that aim at improving image reconstruction and synthesis quality.
Specifically, we select the widely used spatial-based method, \ie, perceptual loss (PL)~\cite{perceptualloss}, which depends on high-level features from a pre-trained VGG~\cite{vgg} network.
We also study the frequency-based approach, \ie, spectral regularization (SpReg)~\cite{specreg}, which is derived based on the azimuthal integration of the Fourier power spectrum.
Besides, we further compare with another transformation form for FFL, \ie, discrete cosine transform (DCT).

The comparison results are reported in Table~\ref{tbl:losscompare}.
FFL outperforms the relevant approaches (\ie, PL and SpReg) when applied to our baselines in different image reconstruction and synthesis tasks. It is noteworthy that FFL and PL are complementary, as shown by our previous experiments on SPADE, which also uses PL.
Even if we replace DFT with DCT as the transformation form of FFL, the results are still better than previous methods. The performance is only slightly inferior to that obtained by FFL with DFT (\ie, Eq.~\eqref{eq:10}). We deduce that the transformation form for FFL may be flexible. At this stage, DFT may be a better choice.

%(PL, freq-based, delve into other transformation form (DCT))

%%\vspace{0.1cm}
%\noindent
%\textbf{Variant studies.}
%
%(Based on Sec. 3.4.)

\begin{figure}[t]
	\centering
%	\vspace{-0.1cm}
	\includegraphics[width=\linewidth]{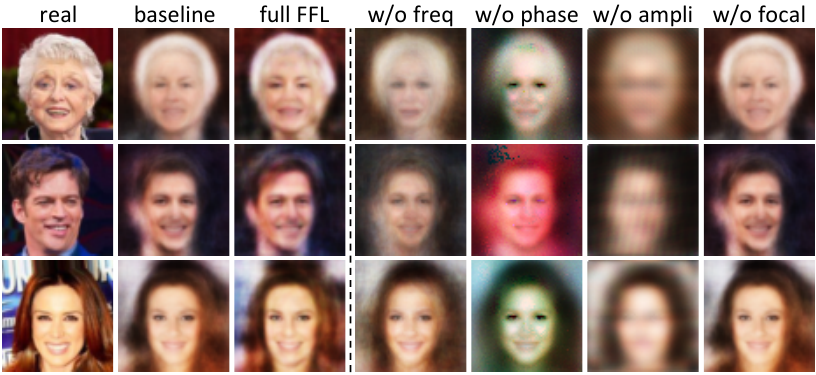}
	\vspace{-0.5cm}
	\caption{\textbf{Ablation studies} of each key component for the focal frequency loss (FFL), \ie, frequency representation (freq), phase and amplitude (ampli) information, and dynamic spectrum weighting (focal) in the vanilla AE image reconstruction task on CelebA.}
	\label{fig:ablation}
	\vspace{-0.1cm}
\end{figure}

\vspace{0.05cm}
\noindent
\textbf{Ablation studies.}
We present ablation studies of each key component for FFL in Figure~\ref{fig:ablation} and corresponding quantitative results in Table~\ref{tbl:ablation}.
For intuitiveness and simplicity, we use vanilla AE image reconstruction on CelebA for the evaluation.

The full FFL achieves the best performance.
%
%Trained with the full FFL, the model achieves better performance than the baseline without FFL.
%
If we do not use the frequency representation of images (Section~\ref{sec:freqrepre}) and focus the model on hard pixels in the spatial domain, the synthesized images become more blurry. The quantitative results degrade.
Discarding either the phase or amplitude information (Section~\ref{sec:freqdist}) harms the metric performance vastly. Visually, using no phase information (amplitude only), the contour of reconstructed faces is retained, but the color is completely shifted. Without amplitude (phase only), the model cannot reconstruct the faces at all, and the full identity information is lost. This further verifies the necessity of considering both amplitude and phase information.
Without focusing the model on the hard frequencies by the dynamic spectrum weighting (\ie, directly using Eq.~\eqref{eq:8}), the results are visually similar to baseline, in line with our discussion in Section~\ref{sec:weightmatrix}. The metrics decrease, being close to but slightly better than baseline, which may benefit from the frequency representation.

\begin{table}[tb!]
%\vspace{-0.1cm}
%\addtolength{\tabcolsep}{3pt}
\centering
\footnotesize
\caption{The PSNR (higher is better), SSIM (higher is better), LPIPS (lower is better), FID (lower is better) and LFD (lower is better) scores for the \textbf{ablation studies} of each key component for the focal frequency loss (FFL).}
%\vspace{0.1cm}
\begin{tabularx}{\linewidth}{c|*{5}{|Y}}
\Xhline{1pt}
 & PSNR$\uparrow$& SSIM$\uparrow$& LPIPS$\downarrow$& FID$\downarrow$& LFD$\downarrow$ \\
\cline{2-6}
%& \multicolumn{5}{c}{DTD} \\
\Xhline{0.6pt}
baseline & 20.044& 0.568& 0.237& 97.035& 14.785 \\
full FFL & {\bf21.703}& {\bf0.642}& {\bf0.199}& {\bf83.801}& {\bf14.403} \\
\Xhline{0.4pt}
w/o freq & 18.200& 0.470& 0.265& 123.833& 15.210 \\
w/o phase & 13.273& 0.380& 0.407& 233.170& 16.344 \\
w/o ampli & 15.640& 0.359& 0.539& 323.528& 15.799 \\
w/o focal & 20.163& 0.574& 0.234& 94.497& 14.758 \\
\Xhline{1pt}
\end{tabularx}
\label{tbl:ablation}
\vspace{-0.55cm}
\end{table}

%(focal image (Sec.3.1), amplitude/phase (as mentioned in Sec.3.2), w/wo focal (as mentioned in Sec.3.3, verify here), variants (supp): alpha/beta parameters, patch, weight, stage-wise)

%\subsection{Settings}
%
%%\vspace{0.1cm}
%\noindent
%\textbf{Baselines.}
%
%(Vanilla AE (2-layer MLP), VAE, pix2pix, SPADE.)
%
%\vspace{0.1cm}
%\noindent
%\textbf{Implementation details.}
%
%(check the ReproducibilityChecklist.pdf)
%
%\vspace{0.1cm}
%\noindent
%\textbf{Datasets.}
%
%(At least 7 datasets. Describe the details here.)
%
%\vspace{0.1cm}
%\noindent
%\textbf{Evaluation metrics (can merge with baselines).}
%
%(1) AE and VAE reconstruction: PSNR, SSIM, LPIPS, FID, LFMSE.
%
%(2) Pix2pix and VAE generation: FID, IS.
%
%(3) SPADE: mIoU, Accuracy, FID.
%
%
%\subsection{Results and Analysis}
%
%%\vspace{0.1cm}
%\noindent
%\textbf{Vanilla AE.}
%
%(qualitative examples, quantitative, training loss, \etc)
%
%\vspace{0.1cm}
%\noindent
%\textbf{VAE.}
%
%\vspace{0.1cm}
%\noindent
%\textbf{Pix2pix.}
%
%\vspace{0.1cm}
%\noindent
%\textbf{SPADE.}
%
%\vspace{0.1cm}
%\noindent
%\textbf{Variant studies.}
%
%(Based on Sec. 3.4.)
%
%\vspace{0.1cm}
%\noindent
%\textbf{Ablation studies.}
%
%(w/wo focal (as mentioned in Sec.3.3, verify here), change image-level loss, alpha/beta parameters, weight, stage-wise, may merge with variant studies)
%
%

% !TEX root = ../main_arxiv.tex

\section{Conclusion}
\label{sec:discussion}

The proposed focal frequency loss directly optimizes image reconstruction and synthesis methods in the frequency domain.
The loss adaptively focuses the model on the frequency components that are hard to deal with to ameliorate quality.
The loss is complementary to existing spatial losses of diverse baselines varying in categories, network structures, and tasks, outperforming relevant approaches.
We further show the potential of focal frequency loss to improve synthesis results of StyleGAN2.
Exploring other applications and devising better frequency domain optimization strategies can be interesting future works.

\vspace{0.05cm}
\noindent
\textbf{Acknowledgments.} This study is supported under the RIE2020 Industry Alignment Fund – Industry Collaboration Projects (IAF-ICP) Funding Initiative, as well as cash and in-kind contribution from the industry partner(s).

{\small
\bibliographystyle{ieee_fullname}
\bibliography{sections_arxiv/egbib}
}

% !TEX root = ../main_arxiv.tex

\clearpage
\section*{Appendix}
\label{sec:appendix}
This appendix provides supplementary information that is not elaborated in our main paper:
Section~\ref{sec:addimethodillus} shows some additional illustrations to explain our method further.
Section~\ref{sec:implementation} describes the implementation details in our experiments.
Section~\ref{sec:dataset} details our used datasets under diverse settings.
Section~\ref{sec:variants} provides some studies on the variants of focal frequency loss.
Section~\ref{sec:analysis} presents additional results and analysis.

\appendix
%******************************************************
\section{Additional Illustrations of Methodology}
\label{sec:addimethodillus}

\subsection{Spatial Frequency Visualization}
\label{sec:spatialfreqvis}
After applying 2D discrete Fourier transform, an image is converted into its frequency representation and decomposed into orthogonal sine and cosine functions.
The angular frequency of each sine and cosine function is decided by the frequency spectrum coordinate $\left(u,v\right)$.
The spatial frequency manifests as the 2D sinusoidal components in the image.
The spectrum coordinate also represents the angled direction of a specific spatial frequency.
As an intuitive view, we show some examples of the 2D sinusoidal components with specific spatial frequencies in Figure~\ref{fig:spatialfreq}.
It is observed that the angled direction and density (angular frequency) of the waves depend on the spectrum coordinate $\left(u,v\right)$.
Besides, the complex frequency value $F\left(u,v\right)$ can be regarded as the weight for each wave, and the weighted sum corresponds to the whole image in the spatial domain.

\subsection{More Intuitive Illustration}
\label{sec:intuitiveunders}
\begin{figure}[t]
	\centering
%	\vspace{-0.35cm}
	\includegraphics[width=\linewidth]{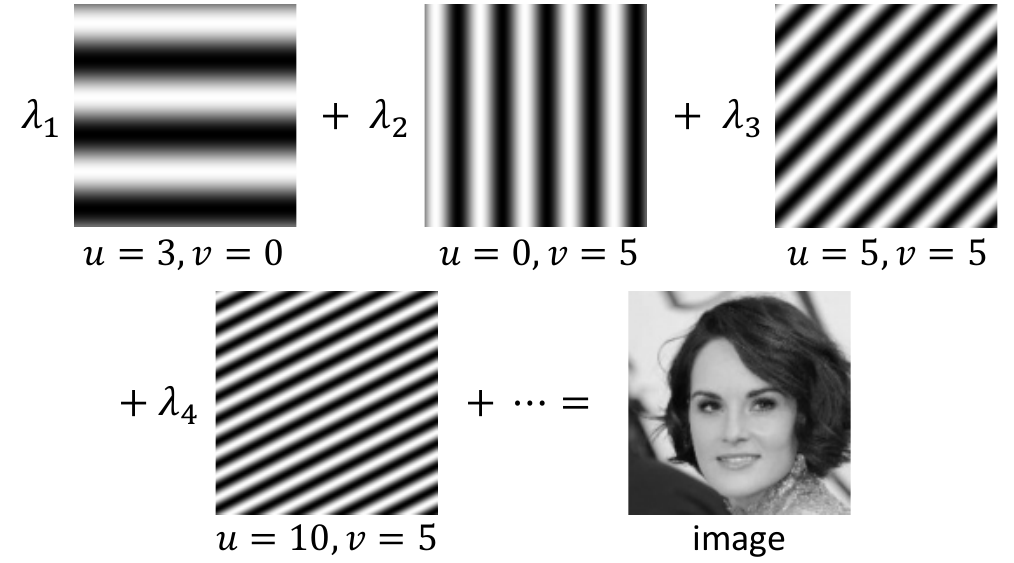}
%	\vspace{-0.5cm}
	\caption{Two-dimensional sinusoidal components with specific spatial frequencies in an image. The angled direction and density (angular frequency) of the waves depend on the spectrum coordinate $\left(u, v\right)$, and $F\left(u,v\right)$ can be seen as the weight for each wave.}
	\label{fig:spatialfreq}
	\vspace{-0.1cm}
\end{figure}

\begin{figure}[t]
	\centering
%	\vspace{-0.35cm}
	\includegraphics[width=0.85\linewidth]{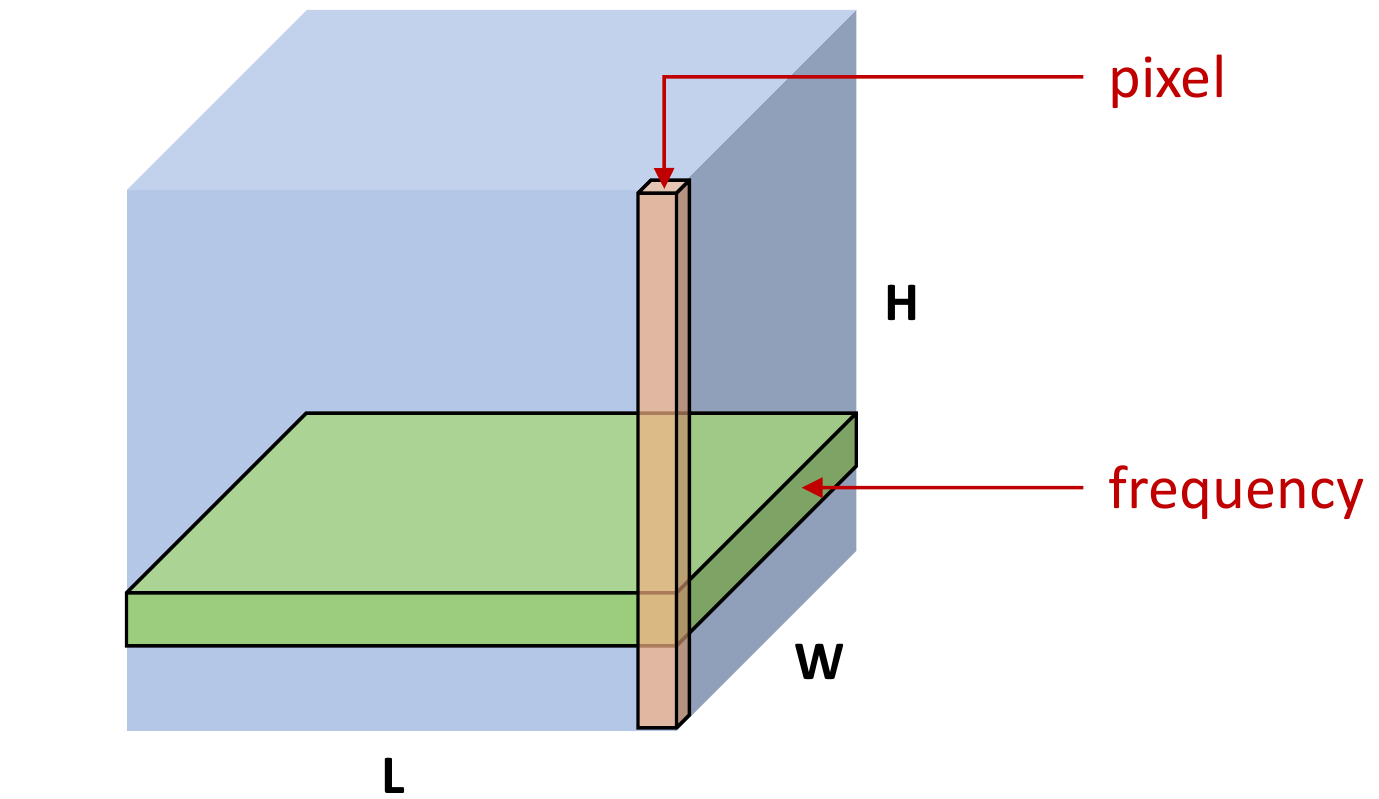}
%	\vspace{-0.5cm}
	\caption{According to Figure~\ref{fig:spatialfreq}, an image (gray-scale for simplicity) can be seen as a cube in a space, where its length (L) and width (W) dimensions correspond to the pixel domain, and the height (H) dimension corresponds to the frequency domain.}
	\label{fig:cubeillustration}
	\vspace{-0.3cm}
\end{figure}

To further explain the proposed focal frequency loss (FFL), we will provide a more intuitive illustration in this section.
According to Figure~\ref{fig:spatialfreq}, an image (gray-scale for simplicity) is the weighted sum of different spatial frequencies. We expand the accumulated frequencies into a new dimension, thus the image can be seen as a cube in a space. The length (L) and width (W) dimensions of the cube correspond to the pixel domain, and the height (H) dimension corresponds to the frequency domain, as shown in Figure~\ref{fig:cubeillustration}. Therefore, a single pixel can be seen as the orange prism, and a specific frequency can be regarded as the green plane.
It is observed that each frequency (\ie, each coordinate value on the frequency spectrum) depends on all the image pixels. Due to the inherent bias of neural networks~\cite{spectralbias,fprinciple}, a model tends to eschew some frequency components that are hard to synthesize, \ie, hard frequencies, in the H dimension. Optimizing in the spatial domain (\ie, in the L and W dimensions) hardly help the model locate these hard frequencies in the H dimension. Similarly, focusing on certain pixels (\eg, orange prism) hardly help the model tackle the hard frequencies (\eg, green plane). Intuitively, when directly optimizing in the H dimension (\ie, explicitly using the frequency representation of the image in our method), the model can easily locate hard frequencies and in turn focus on them.

In Figure~\ref{fig:cubeillustration}, it is noteworthy that each frequency also affects all the image pixels in the spatial domain. When FFL directly optimizes and adaptively focuses a model in the frequency domain, the frequency components in the H dimension will be reconstructed and synthesized better. Meanwhile, the general alignment and quality of all the image pixels in the L and W dimensions will be indirectly improved by FFL, thus boosting some pixel-based metrics (\eg, PSNR and SSIM~\cite{ssim}) and ameliorating the image reconstruction and synthesis quality.

We wish to highlight that both the spatial-based loss and frequency-based loss are important since they consider different aspects and dimensions of an image, as illustrated in Figure~\ref{fig:cubeillustration}. Hence, they are complementary and not replaceable. The proposed FFL is intending to complement existing spatial losses of different methods to improve reconstruction and synthesis quality further.

In fact, the actual situation of the frequency components in an image is much more complicated, which may be a higher-dimensional representation. The visualization in this section just provides a simple and intuitive illustration to help understand the proposed method in this paper.

%******************************************************
\section{Implementation Details}
\label{sec:implementation}
The code used for our experiments will be made publicly available.
All the experiments are conducted on the NVIDIA Tesla V100 GPUs with $32$ GB memory capacity.

\subsection{Baseline Details}
\label{sec:baselinedetails}
In this section, we will provide the implementation details of all the baselines in different image reconstruction and synthesis tasks.
We select five representative methods from the two popular categories: autoencoder-based and GAN-based.
Besides, we evaluate different network structures. Specifically, we explore the multilayer perceptron (MLP) network and the convolutional neural network (CNN). For CNN, the network details also vary, \eg, with or without the skip connections.
In addition, we consider various basic spatial domain losses, \eg, MSE loss, L1 loss, GAN loss~\cite{GAN}, perceptual loss~\cite{perceptualloss}, \etc, to test the ability of focal frequency loss to complement these losses.

%[description/category, task, net (skip con), basic loss]

\vspace{0.05cm}
\noindent
\textbf{Vanilla AE.}
Vanilla autoencoder~\cite{ae} learns the image latent representation in an unsupervised manner, traditionally used for dimension reduction and feature learning.
We employ vanilla AE in the image reconstruction task.
The network is a simple $2$-layer MLP with a hidden size of $256$. We adopt ReLU activations (except the last layer using Tanh) and no norm layers.
We use Adam~\cite{adam} optimizer and set $\beta_1=0.9, \beta_2=0.999$. The learning rate is $0.001$. Normal initialization (with mean $0.0$ and standard deviation $0.02$) is applied to all the networks of vanilla AE.
The spatial loss is MSE loss.
The models are trained on $1$ GPU with a batch size of $128$.
We perform $200$ epochs of training on DTD~\cite{DTD} and $20$ epochs of training on CelebA~\cite{celeba}.

\vspace{0.05cm}
\noindent
\textbf{VAE.}
Exploiting a reparameterization trick, the variational autoencoder~\cite{vae} generates images by learning the latent representation in a probability distribution manner.
We use VAE for image reconstruction and unconditional image synthesis.
We employ CNN for VAE, with typical convolution and transposed convolution layers. Batch normalization~\cite{BN} and Leaky ReLU (with a negative slope of $0.2$, except the last layer using Tanh) are applied. Each convolution layer has a kernel size $4 \times 4$, stride $2$, and zero-padding amount~$1$.
In the encoder, the feature map resolution is halved after each convolution block. Images are down-sampled to $4\times4$, so the number of blocks depends on the input size (\eg, if the input size is $64 \times 64$, there will be $4$ blocks). After an input layer, the number of feature channels is $64$. Then, the number of feature channels will double after each convolution block, while we set a maximum channel number to $512$ to avoid using redundant parameters.
We apply two linear layers to the encoded feature to learn $\mu$ and $\sigma$ for the reparameterization. The latent size is $256$. After reparameterization, an additional linear layer is used to adjust the feature to the original shape.
In the decoder, the network structure is completely inverse to the encoder by replacing convolution layers with the transposed convolution layers.
We use Adam~\cite{adam} optimizer and set $\beta_1=0.9, \beta_2=0.999$. The learning rate is $0.001$. Normal initialization (with mean $0.0$ and standard deviation $0.02$) is applied to all the networks of VAE.
The spatial losses are MSE loss and KL divergence loss~\cite{vae}.
The models are trained on $1$ GPU with a batch size of $128$.
We train our models for $20$ epochs on CelebA~\cite{celeba} and $400$ epochs on CelebA-HQ~\cite{pggan}.

\vspace{0.05cm}
\noindent
\textbf{pix2pix.}
pix2pix~\cite{pix2pix} adopts conditional GAN~\cite{congan} as a general-purpose solution to image-to-image translation with training pairs.
We employ pix2pix for conditional image synthesis.
The U-Net~\cite{unet} generator is applied, which is an encoder-decoder with skip connections between mirrored layers in the encoder and decoder stacks. There are $8$ skip connection blocks in the generator.
The patch-based discriminator is used.
Adam~\cite{adam} optimizer is used with $\beta_1=0.5, \beta_2=0.999$. The learning rate is $0.0002$. Normal initialization (with mean $0.0$ and standard deviation $0.02$) is applied to all the networks of pix2pix.
The spatial losses are vanilla GAN loss~\cite{GAN} and L1 loss.
The models are trained on $1$ GPU.
We conduct $200$ epochs of training on CMP Facades~\cite{cmpfacades} with a batch size of $1$. We train the models for $15$ epochs on edges $\rightarrow$ shoes~\cite{shoesutzappos50K} with a batch size of $4$.
For other detailed network structures and parameters, we follow the original paper~\cite{pix2pix} and their released code.

\vspace{0.05cm}
\noindent
\textbf{SPADE.}
As a task-specific GAN-based method for semantic image synthesis (\ie, synthesizing a photorealistic image from a semantic segmentation mask), SPADE~\cite{SPADE} resizes the segmentation mask for modulating the activations in normalization layers by a learned affine transformation.
The generator is built on a series of residual blocks~\cite{resnet} with the synchronized version of batch normalization. The multi-scale patch-based discriminator~\cite{pix2pixhd} with the instance normalization~\cite{IN} is exploited. Besides, spectral normalization~\cite{spectralnorm} is applied to all the convolutional layers in the generator and discriminator.
Adam~\cite{adam} optimizer is exploited with $\beta_1=0$, $\beta_2=0.9$. Two time-scale update rule~\cite{TTUR} is applied, where the learning rates for the generator and the discriminator are $0.0001$ and $0.0004$, respectively.
The spatial losses are hinge-based GAN loss~\cite{geometricgan,spectralnorm,selfattentiongan}, perceptual loss~\cite{perceptualloss} calculated by VGG-19~\cite{vgg} model, and feature matching loss~\cite{pix2pixhd}.
The models are trained for $200$ epochs on Cityscapes~\cite{cityscapes} and ADE20K~\cite{ade20k} using $4$ GPUs. The batch size is $32$.
For other detailed network structures and parameters, we follow the original paper~\cite{SPADE} and their released code.

\vspace{0.05cm}
\noindent
\textbf{StyleGAN2.}
We further explore the potential of focal frequency loss on the state-of-the-art unconditional image synthesis method, StyleGAN2~\cite{stylegan2}.
We construct the StyleGAN2 baseline on top of its open-source official implementation.
The mapping network consists of $8$ fully connected layers. The dimensionality of both the input latent space and intermediate latent space is $512$.
The activation function is Leaky ReLU with a negative slope of $0.2$.
Several standard techniques in~\cite{pggan,stylegan} are applied, such as the exponential moving average of generator weights, mini-batch standard deviation layer at the end of the discriminator, equalized learning rate for all the trainable parameters, \etc.
Adam~\cite{adam} optimizer is used with $\beta_1=0, \beta_2=0.99$.
The spatial loss is non-saturating logistic loss~\cite{GAN,stylegan2} with $R_1$ regularization~\cite{ganstability}.
All the models are trained with $8$ V100 GPUs. The batch size is $64$ for CelebA-HQ~\cite{pggan} ($256 \times 256$) and $32$ for the resolution of $1024 \times 1024$.
For other detailed network structures and parameters, we follow the original paper~\cite{stylegan2} and their released code.

\vspace{0.05cm}
%\noindent
As for the relevant losses used for comparison, \ie, perceptual loss~\cite{perceptualloss} and spectral regularization~\cite{specreg}, we follow all the details in their papers and released code.

%\vspace{-0.01cm}
\subsection{Computational Cost}
\label{sec:computationalcost}
The computational cost of the proposed focal frequency loss (FFL) is negligible. 
Take pix2pix image-to-image translation on the CMP Facades dataset as an example. The average computational training time only increases from $0.064$ to $0.067$ seconds per iteration after applying FFL.
The memory consumption increases from $3513$ to $3515$ MB.
This cost test is conducted on 1 NVIDIA Tesla V100 GPU.

%******************************************************
\section{Dataset Details}
\label{sec:dataset}
In this section, we will provide detailed information about the seven datasets we explored. The datasets vary in types, sizes, and resolutions.

\begin{itemize}
%	\vspace{-0.15cm}
	\item \textbf{Describable Textures Dataset (DTD).} We use DTD provided by~\cite{DTD}, which is an evolving collection of textural images in the wild, annotated with human-centric attributes. DTD contains texture images with special frequency patterns. We perform vanilla AE image reconstruction using this dataset, with $4,512$ images for training and $1,128$ images for testing. The original images are scaled and center cropped to $64 \times 64$.

	\item \textbf{CelebA.} CelebA~\cite{celeba} is a large-scale face attributes dataset covering large pose variations and background clutter. We conduct image reconstruction with vanilla AE and VAE on CelebA. Besides, we perform VAE unconditional image synthesis on CelebA. We use the cropped and aligned faces, which are more natural images. The training set contains $199,599$ images, and the test set has $3,000$ images. The images are resized and center cropped to $64 \times 64$.

	\item \textbf{CelebA-HQ.} CelebA-HQ is a higher-quality version of the CelebA dataset provided by~\cite{pggan}. The original resolution is $1024 \times 1024$. We perform VAE image reconstruction on this dataset. Besides, we study the unconditional image synthesis by VAE and StyleGAN2 using CelebA-HQ. The dataset is randomly split, yielding $27,000$ images for training and $3,000$ images for evaluation. All the cropped and aligned face images are uniformly resized to $256 \times 256$. For StyleGAN2, we also tried to synthesize images with a resolution of $1024 \times 1024$ besides $256 \times 256$.

	\item \textbf{CMP Facades.} For pix2pix image-to-image translation, we utilize the officially prepared CMP Facades~\cite{cmpfacades} dataset. The facades are collected from different cities around the world with diverse architectural styles. CMP Facades contains architectural labels and photos, which is suitable for mask $\rightarrow$ image translation. The sizes of training and test sets are $400$ and $106$, respectively. The resolution is $256 \times 256$.
	
	\item \textbf{Edges $\rightarrow$ shoes.} We also exploit the officially prepared edges $\rightarrow$ shoes dataset for pix2pix image-to-image translation. The shoe images are from UT Zappos50K~\cite{shoesutzappos50K}. The shoes are centered on a white background. The edge maps are detected by HED~\cite{hededgedet}. The numbers of images for training and testing are $49,825$ and $200$, respectively. The image size is $256 \times 256$.
	
	\item \textbf{Cityscapes.} We use the Cityscapes~\cite{cityscapes} dataset for SPADE semantic image synthesis. Cityscapes dataset consists of street scene images that are mostly collected in Germany. The dataset provides instance-wise, dense pixel annotations of $30$ classes. The training set has $2,975$ images, and the test set contains $500$ images. The images are scaled to $512 \times 256$.

	\item \textbf{ADE20K.} ADE20K~\cite{ade20k} dataset contains challenging in-the-wild images with fine annotations of $150$ semantic classes. We also use ADE20K for SPADE semantic image synthesis, with $20,210$ images for training and $2,000$ images for evaluation. All the images are resized to $256 \times 256$.
%	\vspace{-0.15cm}
\end{itemize}

%******************************************************
\section{Variant Studies}
\label{sec:variants}

In our main paper, we mentioned that the exact form of the proposed focal frequency loss (FFL) is not crucial. In this section, we will provide some variants to extend and modify FFL. We will show some studies on these variants.
For simplicity and intuitiveness, we revisit the vanilla AE image reconstruction task on CelebA.
We report quantitative evaluation results for the variant studies. The visual results of variants are similar.

\begin{table}[tb!]
%\vspace{-0.1cm}
%\addtolength{\tabcolsep}{3pt}
\centering
\footnotesize
\caption{The PSNR (higher is better), SSIM (higher is better), LPIPS (lower is better), FID (lower is better) and LFD (lower is better) scores for the \textbf{variant studies} on the spectrum weight matrix \textbf{parameter} $\bm\alpha$ for the focal frequency loss.}
%\vspace{0.1cm}
\begin{tabularx}{\linewidth}{c|*{5}{|Y}}
\Xhline{1pt}
 & PSNR$\uparrow$& SSIM$\uparrow$& LPIPS$\downarrow$& FID$\downarrow$& LFD$\downarrow$ \\
\cline{2-6}
%& \multicolumn{5}{c}{DTD} \\
\Xhline{0.6pt}
baseline & 20.044& 0.568& 0.237& 97.035& 14.785 \\
\Xhline{0.4pt}
$\alpha=1$ (main) & {\bf21.703}& {\bf0.642}& 0.199& 83.801& {\bf14.403} \\
$\alpha=2$ & 21.376& 0.621& 0.203& 102.329& 14.478 \\
$\alpha=0.5$ & 21.521& 0.635& {\bf0.197}& {\bf82.561}& 14.445 \\
$\alpha=0.1$ & 20.497& 0.591& 0.225& 89.792& 14.681 \\
\Xhline{1pt}
\end{tabularx}
\label{tbl:variantsalpha}
%\vspace{-0.57cm}
\end{table}

\begin{table}[tb!]
%\vspace{-0.1cm}
%\addtolength{\tabcolsep}{3pt}
\centering
\footnotesize
\caption{The PSNR (higher is better), SSIM (higher is better), LPIPS (lower is better), FID (lower is better) and LFD (lower is better) scores for the \textbf{variant studies} on \textbf{patch-based} focal frequency loss. Patch factor $p$ is the number of patches on each edge.}
%\vspace{0.1cm}
\begin{tabularx}{\linewidth}{c|*{5}{|Y}}
\Xhline{1pt}
 & PSNR$\uparrow$& SSIM$\uparrow$& LPIPS$\downarrow$& FID$\downarrow$& LFD$\downarrow$ \\
\cline{2-6}
%& \multicolumn{5}{c}{DTD} \\
\Xhline{0.6pt}
baseline & 20.044& 0.568& 0.237& 97.035& 14.785 \\
\Xhline{0.4pt}
$p=1$ (main) & 21.703& 0.642& 0.199& {\bf83.801}& 14.403 \\
$p=2$ & {\bf21.836}& {\bf0.648}& 0.185& 88.475& {\bf14.372} \\
$p=4$ & 21.752& 0.643& {\bf0.170}& 90.612& 14.392 \\
$p=8$ & 21.414& 0.627& 0.176& 102.334& 14.470 \\
\Xhline{1pt}
\end{tabularx}
\label{tbl:variantspatch}
\vspace{-0.2cm}
\end{table}

Several simple variants can be derived by adjusting the spectrum weight matrix parameter $\alpha$. The parameter $\alpha$ controls how close the weight matrix values are, \ie, how focused the model is. The larger $\alpha$ is, the model will be more focused on the hard frequencies, \ie, the weight difference for easy and hard frequencies will be larger. For the experiments we present in our main paper, we set $\alpha=1$ (we call the main version).
The results are shown in Table~\ref{tbl:variantsalpha}.
Applying the main version of FFL ($\alpha=1$) shows better performance than the baseline without FFL in all the five metrics.
If we set $\alpha=2$, the quantitative results degrade from the main version, especially FID. This suggests that the model may be too focused on the hard frequencies while ignoring some important easy frequency information, albeit the results are still better than the baseline in most cases.
When setting $\alpha=0.5$, all the metric results are better than the baseline. The LPIPS and FID scores become better than the main version. The results of this variant are close to the main version of FFL.
If we set $\alpha=0.1$, the quantitative results degrade from the main version despite still better than the baseline. This indicates that the model may be too unfocused.
For a trade-off, we select $\alpha=1$ as the main version of FFL, while one may consider choosing other variants regarding the parameter $\alpha$ in certain tasks for the flexibility.

Besides, we study another category of variants, the patch-based focal frequency loss, where we crop an image into small patches so that the focused frequencies are at the patch level.
We define the patch factor $p$ as the number of patches on each edge. For instance, if $p=2$, the image will be cropped into $2 \times 2=4$ patches. Obviously, using the original image without cropping it into patches, \ie, the main version of FFL we defined before, corresponds to $p=1$.
The results are shown in Table~\ref{tbl:variantspatch}.
We note that $p=1,2,4$ achieve close performance regarding the five evaluation metrics, all of which are much better than the baseline.
However, if we set $p=8$, the quantitative performance will degrade from the previous versions, especially FID. Although the results are still better than the baseline in most cases, this indicates that the patch size should not be too small.
We simply choose $p=1$ as the main version of FFL for our experiments in the main paper.
However, the variant studies show that the patch-based focal frequency loss may contribute to an additional performance boost in certain cases.
Thus, this may be another direction to extend and modify FFL.

%******************************************************
\section{Additional Results and Analysis}
\label{sec:analysis}

\begin{figure}[t]
	\centering
%	\vspace{-0.35cm}
	\includegraphics[width=\linewidth]{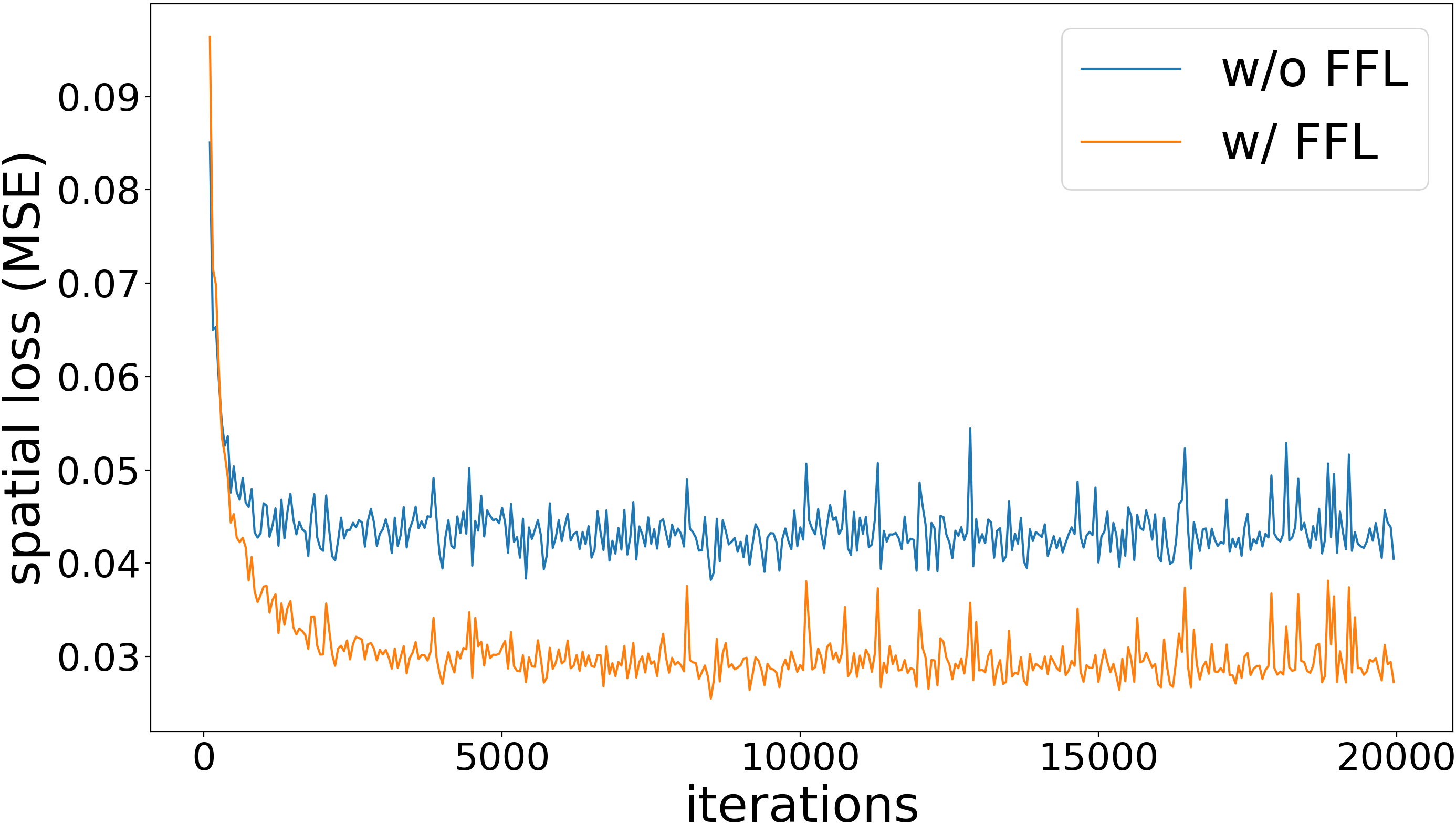}
%	\vspace{-0.5cm}
	\caption{The spatial losses (MSE) with the same weight and random seed of the two training processes with/without focal frequency loss (FFL) for vanilla AE image reconstruction on CelebA. The spatial loss converges to a lower point with the help of FFL.}
	\label{fig:trainloss}
	\vspace{-0.2cm}
\end{figure}

\subsection{Training Loss}
\label{sec:trainingloss}
In the main paper, we have mentioned that the proposed focal frequency loss (FFL) is complementary to existing spatial losses, \eg, MSE loss, to improve image reconstruction and synthesis quality.
We further analyze the training loss in this section. We choose the vanilla AE image reconstruction task on CelebA~\cite{celeba} for simplicity.
We plot the spatial losses with the same weight and random seed of the two training processes with/without FFL in Figure~\ref{fig:trainloss}.
It is readily observed that the spatial loss (MSE) converges to a lower point after applying FFL.
This indicates that the model may converge to a better point with the help of FFL, in line with the better perceptual quality and quantitative performance we presented in our main paper.

\subsection{Frequency Domain Gap}
\label{sec:freqgap}

\begin{figure*}[t]
	\centering
%	\vspace{-0.35cm}
	\includegraphics[width=\linewidth]{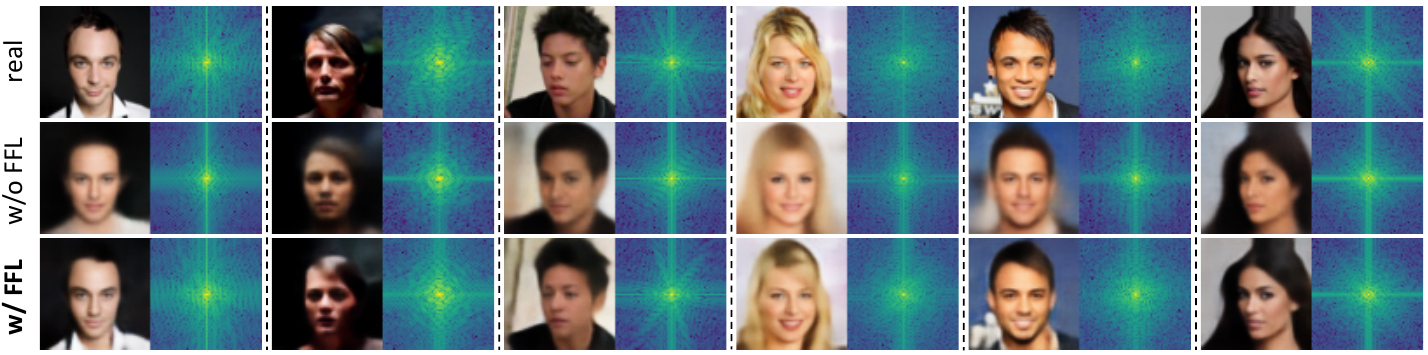}
%	\vspace{-0.5cm}
	\caption{Frequency domain gaps are narrowed by the focal frequency loss (FFL) for VAE image reconstruction on CelebA.}
	\label{fig:freqgap}
%	\vspace{-0.25cm}
\end{figure*}

As mentioned in the main paper, we wish to improve the image reconstruction and synthesis quality by narrowing the frequency domain gap between the real and generated images using the proposed focal frequency loss (FFL).
We have shown that the gaps between mini-batch average spectra of state-of-the-art StyleGAN2 are clearly mitigated by FFL.
We will show some more examples of VAE image reconstruction on the CelebA~\cite{celeba} dataset and provide more analysis about the frequency domain gap in this section.

The results are shown in Figure~\ref{fig:freqgap}.
In the spatial domain, without applying FFL, the reconstructed faces are blurry. This may be attributed to the reparameterization operation in the latent space between the encoder and decoder, which increases the difficulty for reconstruction.
Trained with FFL, the VAE model can synthesize much clearer results, being closer to the ground truth real images. The perceptual quality is better after applying FFL.
In the frequency domain, in line with our visualizations in the main paper, the VAE baseline without FFL bias to a limited spectrum region, losing high-frequency information (outer regions and corners).
The frequency domain gaps are clearly narrowed after adopting FFL. The spectrum distribution becomes closer to the ground truth. Besides, some essential special spectrum patterns can be generated by applying FFL.
This suggests the effectiveness of focal frequency loss to narrow the frequency domain gaps and ameliorate image quality further.

\subsection{Additional Ablation Studies}
\label{sec:addiablation}

\begin{figure}[t]
	\centering
%	\vspace{-0.1cm}
	\includegraphics[width=\linewidth]{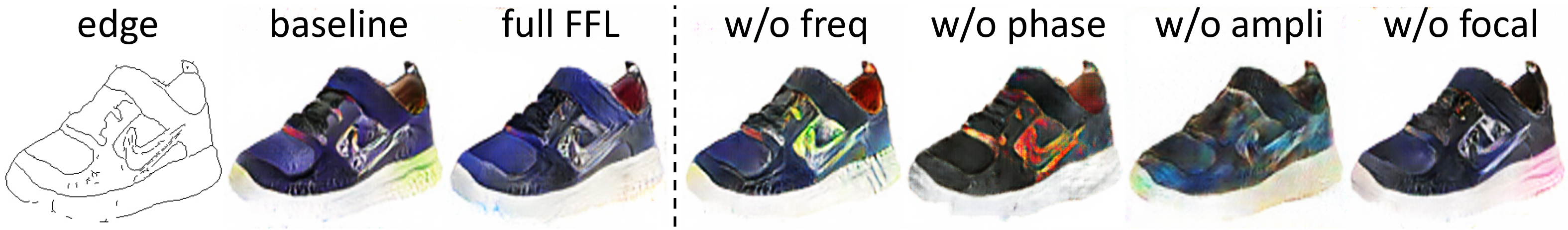}
%	\vspace{-0.5cm}
	\footnotesize
	\begin{tabularx}{\linewidth}{l|*{6}{|Y}}
	%\Xhline{0.8pt}
	\cline{1-7}
	%Metric& baseline& full FFL& w/o freq& w/o phase& w/o ampli& w/o focal  \\
	%\Xhline{0.4pt}
	FID $\downarrow$ & 80.279& {\bf74.359}& 86.674& 98.778& 89.255& 77.864 \\
	%\Xhline{0.4pt}
	IS $\uparrow$ & 2.674& {\bf2.804}& 2.713& 2.667& 2.527& 2.705 \\
	%\Xhline{0.8pt}
	\cline{1-7}
	\end{tabularx}
	\vspace{0.05cm}
	\caption{\textbf{Additional ablation studies} of each key component for the focal frequency loss (FFL), \ie, frequency representation (freq), phase and amplitude (ampli) information, and dynamic spectrum weighting (focal) in the pix2pix image-to-image translation task on edges $\rightarrow$ shoes ($256 \times 256$). The corresponding FID (lower is better) and IS (higher is better) scores are reported below the images.}
	\label{fig:addiablation}
	\vspace{-0.2cm}
\end{figure}

In the main paper, we provided the ablation studies of vanilla AE image reconstruction on CelebA for intuitiveness and simplicity, intending to study the importance of each key component for the proposed focal frequency loss (FFL) while reducing the influence of other factors, such as the adversarial loss.
In this section, we provide the additional ablation studies on higher-resolution images with GAN. We show the studies of pix2pix~\cite{pix2pix} (\ie, GAN-based method) image-to-image translation on edges $\rightarrow$ shoes ($256 \times 256$) in Figure~\ref{fig:addiablation}. The results are in line with the ablation studies in our main paper, further suggesting the importance of each key component for FFL.

%\begin{table}[t]
%\centering
%\footnotesize
%\vspace{-0.75cm}
%\begin{tabularx}{\linewidth}{l|*{6}{|Y}}
%%\Xhline{0.8pt}
%\cline{1-7}
%%Metric& baseline& full FFL& w/o freq& w/o phase& w/o ampli& w/o focal  \\
%%\Xhline{0.4pt}
%FID $\downarrow$ & 80.279& {\bf74.359}& 86.674& 98.778& 89.255& 77.864 \\
%%\Xhline{0.4pt}
%IS $\uparrow$ & 2.674& {\bf2.804}& 2.713& 2.667& 2.527& 2.705 \\
%%\Xhline{0.8pt}
%\cline{1-7}
%\end{tabularx}
%\vspace{-0.45cm}
%\end{table}

\subsection{Results on Non-Photorealistic Images}
\label{sec:nonphotorealistic}

We further study the benefit of the proposed focal frequency loss (FFL) on non-photorealistic images.
As an example, we provide the vanilla AE image reconstruction results on Danbooru2019 Portraits~\cite{danbooru2019portraits} (Anime) in Table~\ref{tbl:nonphotorealistic}. Empirically, we observe that all the metrics can still be boosted by FFL.
Our intuition is that FFL can also help generate non-photorealistic images since they still possess special frequency patterns that may be hard for a network to learn. FFL is adaptive for dealing with these frequencies.

\begin{table}[t]
\centering
\footnotesize
%\vspace{-0.31cm}
\caption{The PSNR (higher is better), SSIM (higher is better), LPIPS (lower is better), FID (lower is better) and LFD (lower is better) scores for the \textbf{vanilla AE image reconstruction} on \textbf{Danbooru2019 Portraits (Anime)} trained with/without the focal frequency loss (FFL).}
\begin{tabularx}{\linewidth}{c|c|*{5}{|Y}}
%\Xhline{0.8pt}
\Xhline{1pt}
Dataset& FFL & PSNR$\uparrow$& SSIM$\uparrow$& LPIPS$\downarrow$& FID$\downarrow$& LFD$\downarrow$ \\
%\cline{2-7}
%& \multicolumn{5}{c}{DTD} \\
%\Xhline{0.4pt}
\Xhline{0.6pt}
Anime& w/o & 19.885& 0.575& 0.294& 193.342& 14.822 \\
($64 \times 64$) & w/ & {\bf20.657}& {\bf0.628}& {\bf0.267}& {\bf184.443}& {\bf14.644} \\
%\Xhline{0.8pt}
\Xhline{1pt}
\end{tabularx}
\label{tbl:nonphotorealistic}
\vspace{-0.2cm}
\end{table}

\subsection{Higher-Resolution Results on StyleGAN2}
\label{sec:moreexamples}

\begin{figure*}[t]
	\centering
%	\vspace{-0.15cm}
	\includegraphics[width=\linewidth]{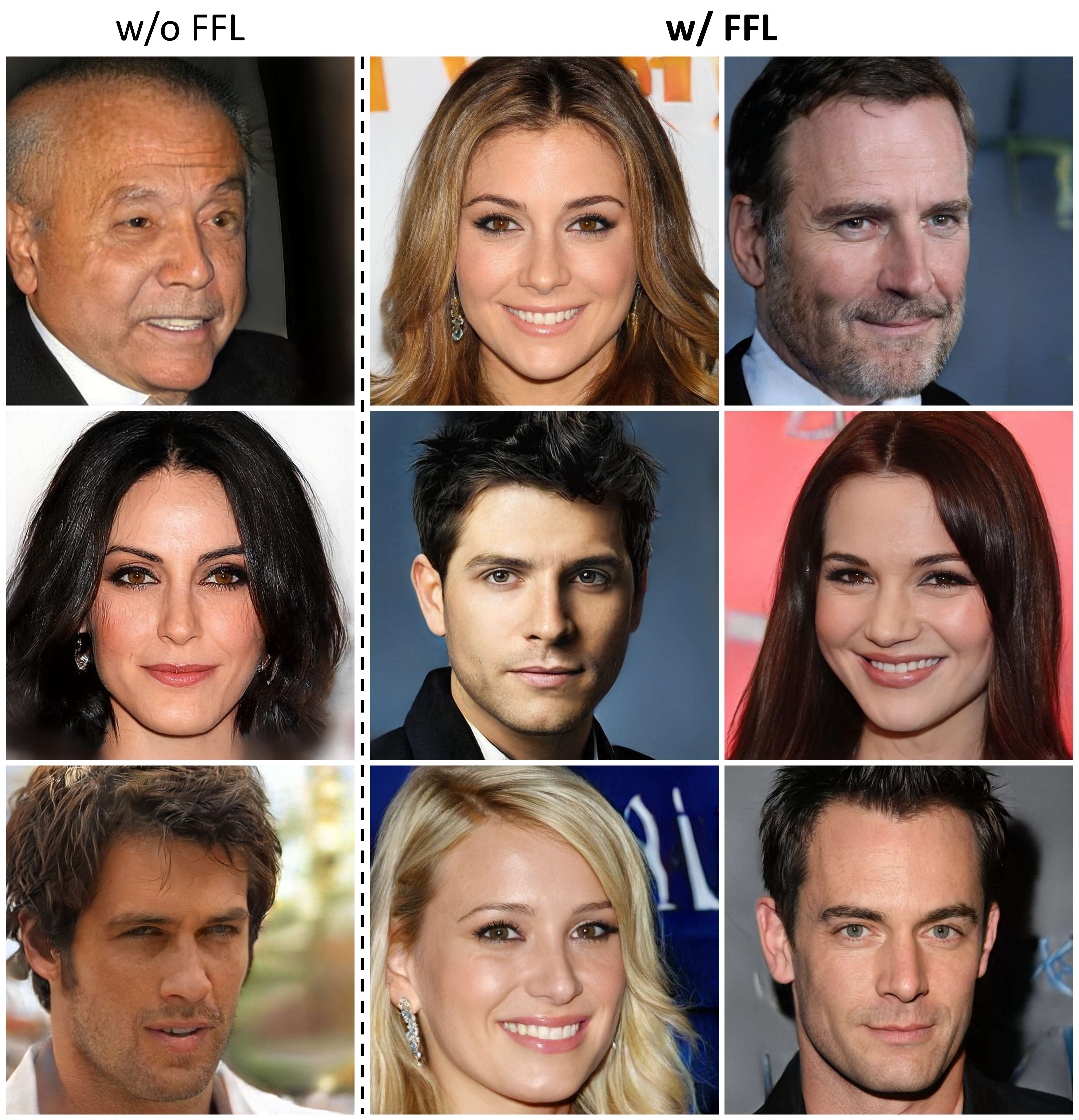}
%	\vspace{-0.05cm}
	\caption{Synthesis results (without truncation) of StyleGAN2 trained with/without the proposed FFL on CelebA-HQ ($1024 \times 1024$). The model with FFL achieves the FID score of $\bf3.374$, outperforming the original StyleGAN2 without FFL of $3.733$.}
	\label{fig:sg2hq1024six}
%	\vspace{-0.18cm}
\end{figure*}

\begin{figure*}[t]
	\centering
%	\vspace{-0.15cm}
	\includegraphics[width=\linewidth]{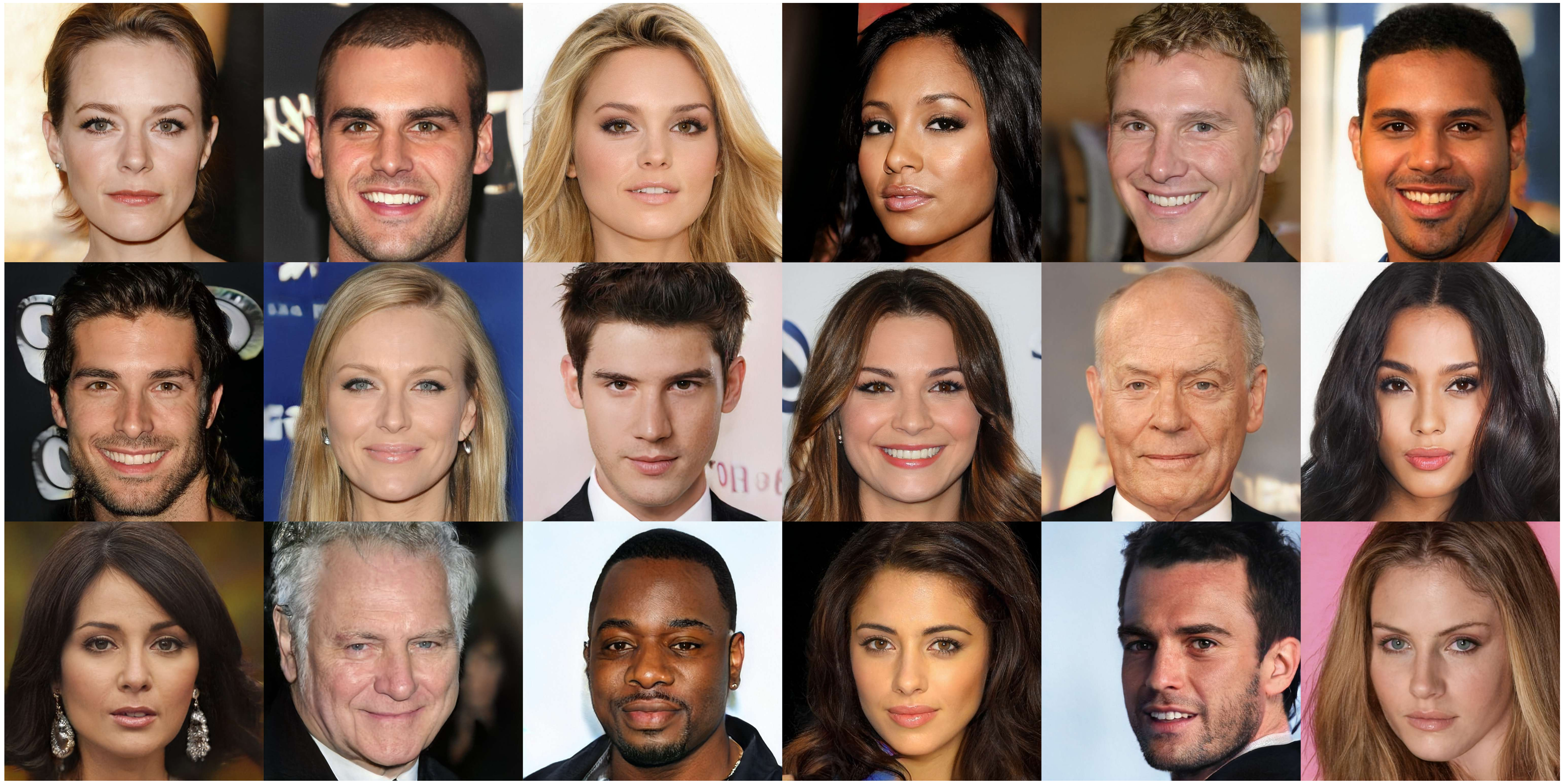}
%	\vspace{-0.05cm}
	\caption{More random sampled images (without truncation) synthesized by StyleGAN2 trained with the proposed FFL on CelebA-HQ ($1024 \times 1024$).}
	\label{fig:sg2hq1024more}
%	\vspace{-0.18cm}
\end{figure*}

\begin{figure*}[t]
	\centering
%	\vspace{-0.15cm}
	\includegraphics[width=\linewidth]{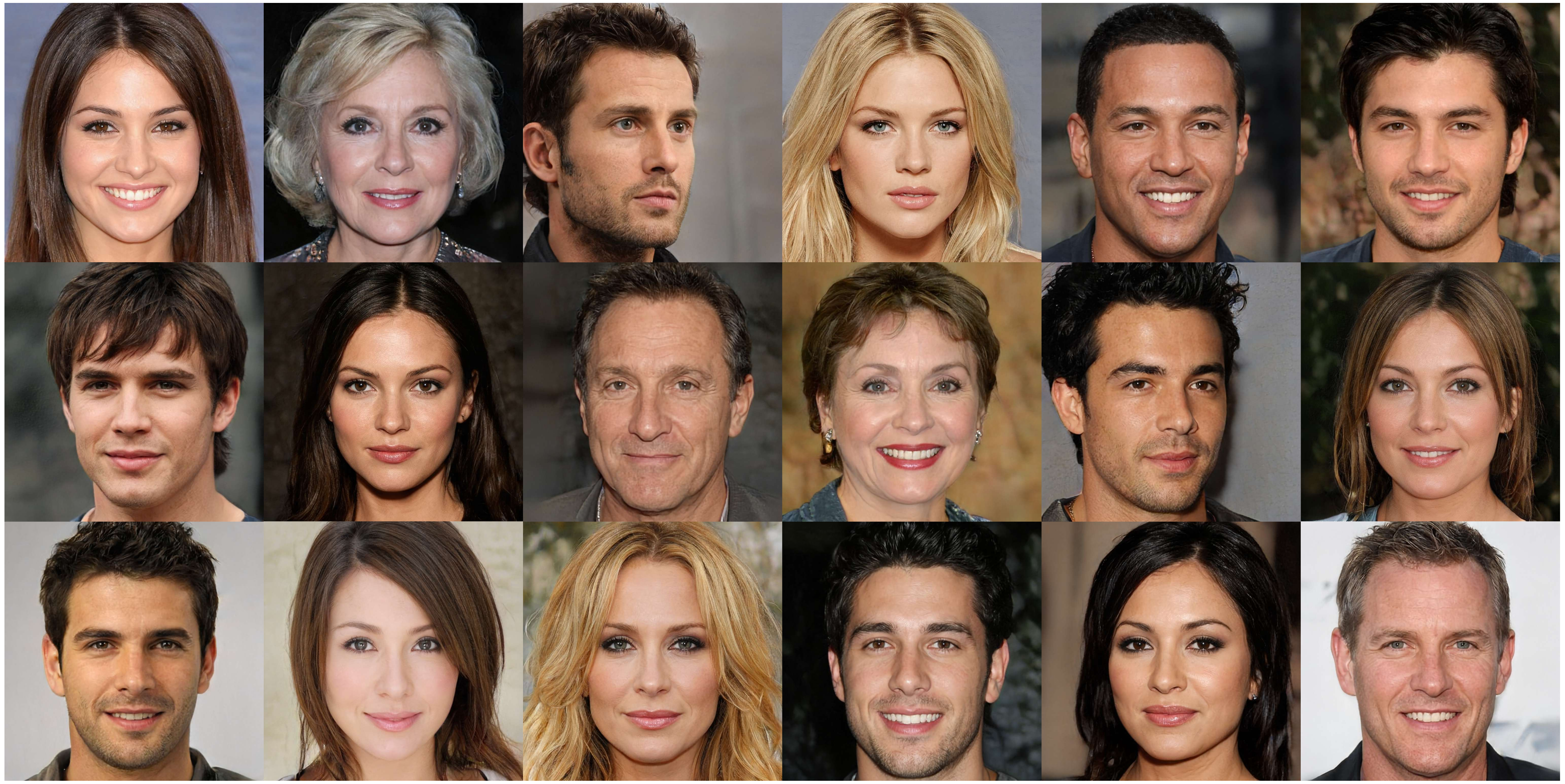}
%	\vspace{-0.05cm}
	\caption{More random sampled images (with truncation applied using $\psi=0.5$~\cite{stylegan,stylegan2}) synthesized by StyleGAN2 trained with the proposed FFL on CelebA-HQ ($1024 \times 1024$).}
	\label{fig:sg2hq1024trunc05}
%	\vspace{-0.18cm}
\end{figure*}

In Figure~\ref{fig:sg2hq1024six}, we show some higher-resolution images synthesized by StyleGAN2~\cite{stylegan2} trained with or without the proposed focal frequency loss (FFL) on CelebA-HQ ($1024 \times 1024$). The truncation trick~\cite{stylegan,stylegan2} is not applied.
Although the original StyleGAN2 (w/o FFL) generates plausible images in most cases, it sometimes produces tiny artifacts on the face (Row $2$) and eyes (Row $3$). The details on the teeth are missing in certain cases (Row $1$). The synthesized images by StyleGAN2 with FFL (w/ FFL) are very photorealistic. Besides, StyleGAN2 achieves a better FID score after applying FFL, indicating that the quality of generated images becomes better with the help of FFL.
More random sampled synthesized images without truncation are shown in Figure~\ref{fig:sg2hq1024more}, and the examples with truncation using $\psi=0.5$~\cite{stylegan,stylegan2} are presented in Figure~\ref{fig:sg2hq1024trunc05}. It is observed that all the images generated by StyleGAN2 with FFL are with very high fidelity.

\end{document}